\documentclass[11pt]{article}
\usepackage{acl}
\usepackage{times}
\usepackage{latexsym}
\usepackage[T1]{fontenc}
\usepackage[utf8]{inputenc}
\usepackage{microtype}
\usepackage{booktabs}
\usepackage{multirow}
\usepackage{amsmath,amssymb}
\usepackage{graphicx}
\usepackage{xr-hyper}
\usepackage{orcidlink}
\usepackage{hyperref}
\usepackage{xcolor}
\usepackage{colortbl}
\usepackage{pifont}
\usepackage{tikz}
\usepackage{algorithm}
\usepackage{algpseudocode}
\usetikzlibrary{arrows.meta,positioning}
\newcommand{\cmark}{\textcolor{teal!70!black}{\ding{51}}}
\newcommand{\xmark}{\textcolor{red!70!black}{\ding{55}}}
\definecolor{crowhi}{HTML}{F0F0F0}
\definecolor{cident}{HTML}{8FBBD9}
\definecolor{cpred}{HTML}{7BC8A4}
\definecolor{caggr}{HTML}{F4B247}
\definecolor{cunder}{HTML}{D98585}
\definecolor{cunable}{HTML}{B9B9B9}
\title{The Warrant Gap:\\Claim-Conditioned Re-scoring for Fact-Checking}

\author{
Arka Ujjal Dey~\orcidlink{0000-0001-8392-1574} 
and John Collomosse~\orcidlink{0000-0003-3580-4685}
}

\begin{document}
\maketitle

\begin{abstract}
Fact-checking systems built on LLMs achieve high verdict accuracy on standard benchmarks, yet routinely output \textsc{Supports} labels whose cited evidence does not license the claim. Structured decomposition is the natural way to inspect those warrants, but rigid extraction protocols strip the full-claim context that facets need. We introduce \textbf{SIFT}---claim-conditioned re-scoring of extracted evidence spans against the full claim---paired with \textbf{WSP} (Warranted Supports Proportion), an automatic NLI check that the cited warrant entails the claim. We evaluate on FEVER, SciFact, 5PILS, and DP across four open-source backbones. SIFT recovers accuracy on cells where naive decomposition costs up to 27.6 points, while raising WSP above direct prompting; WSP itself calibrates against human gold evidence at AUC $0.92$ and precision $0.98$ ($n=2036$).
\end{abstract}

\section{Introduction}

\begin{table*}[t]
\centering
\scriptsize
\setlength{\tabcolsep}{4pt}
\begin{tabular}{@{}p{3.0cm}p{4.3cm}p{4.3cm}p{3.6cm}@{}}
\toprule
\textbf{Shortcut} & \textbf{Why it looks plausible} & \textbf{Where it fails} & \textbf{Diagnostic / repair} \\
\midrule
\textbf{Topical overlap} & The warrant shares named entities or salient words with the claim. & The cited span is about the right topic, but a different predicate, quantity, date, or event. & WSP; claim-conditioned re-scoring. \\
\textbf{Local facet coverage} & Each active 5W1H slot has some supporting span. & The spans license nearby propositions, not the full claim after the facets are recombined. & SIFT: re-score extracted spans against the full claim. \\
\textbf{Patchwork aggregation} & Each document supports only one claim facet. & The documents do not jointly describe one coherent state of affairs. & Coverage matrix plus Late-stage verifier. \\
\textbf{Warrant-safe abstention} & A system avoids inadmissible \textsc{Supports} by predicting non-support. & Accuracy and recall fall, even though WSP can look high. & Report accuracy and WSP together. \\
\bottomrule
\end{tabular}
\caption{\textbf{Fallible warrant shortcuts.} We make explicit the shortcuts that can yield benchmark-correct \textsc{Supports} verdicts without an admissible warrant. The method targets different shortcuts at different stages rather than treating accuracy as a sufficient test of reasoning.}
\label{tab:warrant-shortcuts}
\vspace{-15pt}
\end{table*}

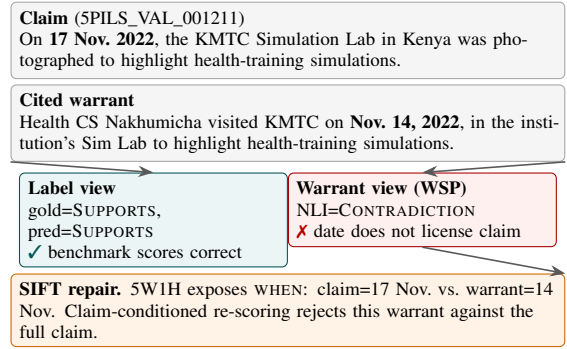
\begin{figure}[t]
\centering
\scriptsize
\begin{tikzpicture}[
  font=\scriptsize,
  >=Stealth,
  box/.style={draw=black!35, rounded corners=2pt, align=left, inner sep=3.2pt},
  claim/.style={box, fill=gray!8, text width=.92\linewidth},
  good/.style={box, fill=teal!8, draw=teal!65!black, text width=.43\linewidth},
  bad/.style={box, fill=red!6, draw=red!65!black, text width=.43\linewidth},
  repair/.style={box, fill=orange!10, draw=orange!80!black, text width=.92\linewidth},
  arrow/.style={->, semithick, draw=black!65}
]
\node[claim] (c) {\textbf{Claim} (5PILS\_VAL\_001211)\\
On \textbf{17 Nov.\ 2022}, the KMTC Simulation Lab in Kenya was photographed to highlight health-training simulations.};
\node[claim, below=2pt of c] (w) {\textbf{Cited warrant}\\
Health CS Nakhumicha visited KMTC on \textbf{Nov.\ 14, 2022}, in the institution's Sim Lab to highlight health-training simulations.};
\node[good, below left=4pt and 0pt of w.south] (acc) {\textbf{Label view}\\
gold=\textsc{Supports}, pred=\textsc{Supports}\\
\cmark\ benchmark scores correct};
\node[bad, below right=4pt and 0pt of w.south] (vsp) {\textbf{Warrant view (WSP)}\\
NLI=\textsc{Contradiction}\\
\xmark\ date does not license claim};
\node[repair, below=42pt of w] (r) {\textbf{SIFT repair.} 5W1H exposes \textsc{when}: claim=17 Nov.\ vs.\ warrant=14 Nov. Claim-conditioned re-scoring rejects this warrant against the full claim.};
\draw[arrow] (w.south west) -- (acc.north);
\draw[arrow] (w.south east) -- (vsp.north);
\draw[arrow] (vsp.south) -- (r.north east);
\end{tikzpicture}
\caption{The warrant gap on a real 5PILS item. Label accuracy scores the prediction correct; the cited span licenses a different date.}
\label{fig:warrant-gap}
\vspace{-20pt}
\end{figure}

Fact-checking benchmarks reward verdict correctness \citep{thorne2018fever,wadden2020scifact}, yet a correct verdict is not always a licensed one. A \textsc{Supports} label can be right for the wrong reasons: the cited evidence may be merely topical, partially aligned with the claim, or assembled from incompatible documents (Fig.~\ref{fig:warrant-gap}). We call this gap between label correctness and evidential licensing the \emph{warrant gap}. Standard accuracy cannot see this gap, and surfacing it requires an automatic, claim-level measure of warrant admissibility. Such a measure must catch the specific shortcuts modern fact-checking exhibits: warrant-level patterns such as topical overlap and patchwork aggregation (Table~\ref{tab:warrant-shortcuts}), which do not surface in verdict labels or free-text rationales alone.

The natural fix is to add structure to the rationale: decompose the claim into facets---5W1H (Who, What, When, Where, Why, How)~\citep{urbani2020verifying,tonglet2024image,tonglet2025cove}---and ask whether the evidence covers each one. But decomposition suffers when local facets are scored independently; they lose the semantic context supplied by the full claim---the \emph{decontextualisation} problem in standalone interpretation \citep{choi2021decontextualization}. Thus, while structure is needed for admissibility, naive structure damages accuracy. We locate this burden specifically in \emph{rigid} slot-filling rather than in decomposition itself \citep{hu2025decomp} and supply a targeted repair. \textbf{SIFT} (Structured Inference with Facet-level Tracing) extracts verbatim 5W1H spans, then re-evaluates them against the \emph{full} claim and revises the verdict, restoring the claim-level context decomposition removes (Table~\ref{tab:decomp-repair}).

Measuring whether repaired verdicts actually license their claims requires a single number. \textbf{WSP} (Warranted Supports Proportion) is the fraction of predicted \textsc{Supports} whose cited warrant entails the claim under an NLI scorer. It exposes movement along the accuracy--admissibility frontier---a system can raise accuracy by emitting unsupported \textsc{Supports}, or raise warrant precision by abstaining---so we report both axes, corroborated by independent grounding models and blind LLM audits. SIFT generates a context-aware verdict but struggles to filter out fragile, multi-document aggregations. Design B solves this by deploying a deterministic layer that cross-references the SIFT verdict with independent evidence windows, allowing the system to confidently veto unsupported claims or rescue missing ones. Its scope is bounded by design: a verifier can keep, demote, or rescue, but cannot reconstruct an entity or predicate the extractor never selected (extraction-bound). Residual errors, therefore, shift from aggregation (verifier-fixable) toward identity and predicate failures (extraction-bound), localising the next research frontier on warrant \emph{construction}.

\paragraph{Contributions.}
(i)~We formalise the \emph{warrant gap}---the slack between label correctness and evidential licensing---and operationalise it with WSP, an automatic admissibility diagnostic. WSP calibrates against human gold evidence at AUC~$0.92$ ($n{=}2036$) and its rankings remain stable across NLI families ($\rho{=}0.89$).
(ii)~We locate the cost of structured decomposition in \emph{rigid slotting}, not decomposition itself, and propose \textbf{SIFT} that repairs this through \emph{claim-conditioned re-scoring}---re-evaluating extracted spans against the full claim before the verdict---and closes most of the gap while preserving admissibility gains.
(iii)~A bounded deterministic verifier over the repaired representation shifts residual errors from aggregation to extraction-bound identity and predicate failures, localising the next frontier on warrant \emph{construction} rather than adjudication.

\section{Related Work}
\label{sec:related}

\paragraph{Fact-checking evaluation.}
Fact checking datasets \cite{abdelnabi2022open,tonglet2025cove,wadden2020scifact,thorne2018fever} mostly focus on evaluating verdict correctness against retrieved evidence. A parallel groundedness literature targets faithfulness of generated text---AIS, FActScore, SAFE, RARR \citep{rashkin2021ais,min2023factscore,wei2024safe,gao2023rarr}---and NLI consistency scoring---TRUE, SummaC, MiniCheck \citep{honovich2022true,laban2022summac,tang2024minicheck}. However these do not isolate the verdict-level object we study: whether a discrete \textsc{Supports} decision is licensed by its cited warrant. Ev2R \citep{akhtar2026ev2r} treats this evidence--verdict alignment as a reference-sensitive problem and warns that verdict-level proxies remain gameable by lexical heuristics. Cluster-based retrieval groups evidence into coherent narratives for a generative judge \citep{crave_arka}, but the judgment is free-text and does not reduce to a quantity comparable across cells. WSP differs by combining an extractive premise with a verdict-level admissibility check, so a system is not rewarded for either unsupported \textsc{Supports} or warrant-safe abstention alone. A complementary attribution literature seeks faithful citation via model internals rather than text, e.g.\ MIRAGE \citep{qi2024mirage}; we take an extractive, text-auditable route because the warrant must remain inspectable end-to-end. To systematically measure these inspectable warrants against known failure modes, evaluation must shift from aggregate leaderboards to targeted diagnostics.

\vspace{-10pt}

\paragraph{Diagnostic evaluation.}
HANS \citep{mccoy2019right} showed that high NLI accuracy can hide reliance on fallible syntactic heuristics such as lexical overlap and subsequence matching. We adopt the same diagnostic stance for fact-checking, but the unit of analysis changes. We do not test whether an NLI model itself has learned a syntactic shortcut; we test whether a fact-checking system's \textsc{Supports} verdict is licensed by its warrant. The relevant shortcuts are therefore warrant-level: topical overlap, local facet coverage, patchwork aggregation, and abstention that looks warrant-safe but loses recall. Exposing and mitigating these specific facet-level shortcuts requires a structural scaffold, which makes claim decomposition a necessary, albeit risky, prerequisite.

\paragraph{Decomposition as scaffold and as failure source.}
Claim decomposition is widely used because it makes verification inspectable \citep{kamoi2023wice,popovic2025jedi}. Recent work shows that aggressive decomposition can strip context and inject semantic noise \citep{hu2025decomp}, and \citet{wanner2025dnd} formalise this as a conflict between decomposition (isolating atomic facts) and decontextualisation (re-injecting full-claim context), proposing context-aware verification as the resolution. Our repair is extractive rather than generative: instead of rewriting subclaims to add context back, we re-score the retrieved 5W1H spans against the full claim, which preserves the typed, fixed-arity coverage matrix that the deterministic downstream verifier and the typed failure taxonomy require---and that a variable-length atomic representation does not supply. We treat this fragmentation risk as our primary target. Naive 5W1H is the failure condition, and our repair, claim-conditioned re-scoring, preserves this structural scaffold while restoring full-claim semantics.
\vspace{-10pt}
\paragraph{Warrants and admissibility.}
The notion of a \emph{warrant} comes from argumentation theory \citep{toulmin1958uses}, where evidence must license a claim rather than merely accompany it. We adopt the operational stance without importing a full argumentation calculus: a warrant is admissible if it is verbatim-grounded in evidence and entails the full claim.
\vspace{-10pt}
\paragraph{LLM adjudication and omission.}
LLM-as-judge methods can read a claim, warrant, and rationale in one pass but are sensitive to prompt and format \citep{seo2025verifying,sharma2024sycophancy}. We include generative adjudication as a substantive comparison (\S\ref{sec:results}). Half-truth and omission-aware verification \citep{tang2025tracer,debating2025radar} target a different failure mode (locally entailing spans defeated by hidden context) that SIFT does not solve; we keep \textsc{Undercutter} as a distinct taxonomy label.

\begin{table}[h]
\centering
\scriptsize
\begin{tabular}{@{}p{2.0cm}p{5.1cm}@{}}
\toprule
\textbf{Label} & \textbf{Meaning} \\
\midrule
\textsc{Admissible} & Warrant licenses the full claim. \\
\textsc{Identity} & Wrong entity, object, person, organisation, or event. \\
\textsc{Predicate} & Right entity, wrong property/relation/action/quantity/time. \\
\textsc{Aggregation} & Facets covered separately; combination is not one coherent event. \\
\textsc{Undercutter} & Locally supportive; defeated by context, missing conditions, or omitted contraries. \\
\textsc{Unable} & Malformed, empty, unlocatable, or not judgeable. \\
\bottomrule
\end{tabular}
\caption{Failure taxonomy for auditing predicted \textsc{Supports} warrants.}
\label{tab:taxonomy}
\vspace{-20pt}
\end{table}

\section{Task and Metrics}
\label{sec:task}

\paragraph{Task.}
Given a claim $c$ and evidence pool $\mathcal{E}=\{e_1,\ldots,e_k\}$, a system predicts $\hat{y}\in\{\mathrm{SUPPORTS},\mathrm{REFUTES},\mathrm{NEI}\}$ and, when it predicts \textsc{Supports}, a warrant $\hat{w}$ that must remain auditable as text. We score binary accuracy with \textsc{Supports} as the positive class.

\paragraph{Admissibility and WSP.}
A predicted \textsc{Supports} verdict is \emph{admissible} iff there exists $W\subseteq\mathcal{E}$ that (i) is verbatim-grounded in the pool and (ii) entails $c$. Admissibility is stricter than label correctness: a prediction can be benchmark-correct yet rest on the wrong entity, a weaker predicate, or a patchwork of incompatible facets. 
Let $S = \{i:\hat{y}_i{=}\mathrm{SUPPORTS}\}$ be the set of asserted supports. We measure it with
\[
\mathrm{WSP}=\frac{\big|\{i \in S : \mathrm{NLI}(\hat{w}_i,c_i){=}\textsc{Entail}\}\big|}{\big|S\big|}.
\]
If a predicted SUPPORTS lacks a recoverable warrant, it contributes only to the denominator, effectively lowering the WSP. This is a diagnostic, not a second accuracy: it separates whether the system chose the right verdict from whether the cited text licenses it. We report both axes everywhere.

\paragraph{Failure taxonomy.}
When a \textsc{Supports} warrant is inadmissible we assign a label from: \textsc{Identity} (wrong entity or event), \textsc{Predicate} (right entity but wrong property, action, quantity, or time), \textsc{Aggregation} (facets supported piecewise but not jointly), \textsc{Undercutter} (locally entailing but defeated by surrounding context), or \textsc{Unable} (malformed or unlocatable). Definitions are in Table~\ref{tab:taxonomy}. The taxonomy is designed to test the mechanism: claim-conditioned re-scoring should reduce facet-level fragmentation and nearby-predicate matches; the late-stage verifier should reduce aggregation; neither is expected to solve identity errors or omission undercutters.

\section{Method}
\label{sec:method}
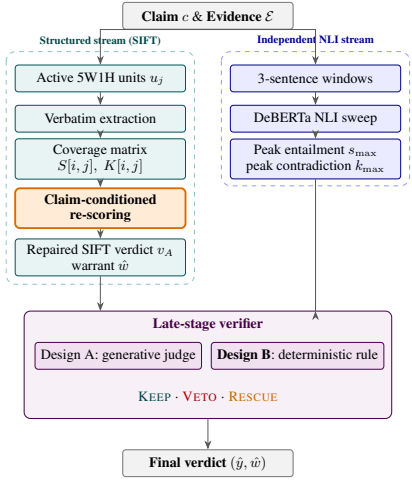
\begin{figure}[t]
\centering
\resizebox{0.7\columnwidth}{!}{%
\begin{tikzpicture}[
  font=\footnotesize,>=Stealth,
  io/.style={draw=black!55,fill=gray!10,rounded corners=2pt,align=center,
             minimum width=3.8cm,minimum height=0.6cm},
  sift/.style={draw=teal!65!black,fill=teal!9,rounded corners=2pt,align=center,
               minimum width=3.8cm,minimum height=0.6cm},
  nli/.style={draw=blue!55!black,fill=blue!8,rounded corners=2pt,align=center,
              minimum width=3.8cm,minimum height=0.6cm},
  hero/.style={draw=orange!85!black,line width=1.2pt,fill=orange!16,
               rounded corners=3pt,align=center,
               minimum width=3.8cm,minimum height=0.7cm},
  vbox/.style={draw=violet!60!black,fill=violet!12,rounded corners=2pt,align=center,
               minimum width=3.6cm,minimum height=0.6cm},
  arr/.style={->,semithick,draw=black!65},
]

\node[io] (in) at (0,0) {\textbf{Claim} $c$ \& \textbf{Evidence} $\mathcal{E}$};

\node[sift] (u)  at (-2.4,-1.4) {Active 5W1H units $u_j$};
\node[sift] (sp) at (-2.4,-2.3) {Verbatim extraction};
\node[sift] (mx) at (-2.4,-3.2) {Coverage matrix\\$S[i,j],\ K[i,j]$};
\node[hero] (rs) at (-2.4,-4.3) {\textbf{Claim-conditioned}\\\textbf{re-scoring}};
\node[sift] (wr) at (-2.4,-5.4) {Repaired SIFT verdict $v_A$\\warrant $\hat{w}$};

\node[nli] (wd) at (2.4,-1.4) {3-sentence windows};
\node[nli] (sw) at (2.4,-2.3) {DeBERTa NLI sweep};
\node[nli] (pk) at (2.4,-3.2) {Peak entailment $s_{\max}$\\peak contradiction $k_{\max}$};

\draw[draw=teal!45,dashed,rounded corners=6pt] (-4.5,-0.8) rectangle (-0.3,-6.0);
\node[font=\scriptsize\bfseries,text=teal!55!black,anchor=south] at (-2.4,-0.8) {Structured stream (SIFT)};

\draw[draw=blue!40,dashed,rounded corners=6pt] (0.3,-0.8) rectangle (4.5,-3.8);
\node[font=\scriptsize\bfseries,text=blue!55!black,anchor=south] at (2.4,-0.8) {Independent NLI stream};

\draw[draw=violet!55!black,fill=violet!6,rounded corners=4pt] (-4.1,-6.6) rectangle (4.1,-9.0);
\node[font=\footnotesize\bfseries,text=violet!55!black] at (0,-6.9) {Late-stage verifier};
\node[vbox] (da) at (-2.0,-7.6) {Design A: generative judge};
\node[vbox] (db) at (2.0,-7.6) {\textbf{Design B}: deterministic rule};
\node[align=center,font=\footnotesize] (oc) at (0,-8.5)
   {\textcolor{teal!55!black}{\textsc{Keep}} $\cdot$ \textcolor{red!65!black}{\textsc{Veto}} $\cdot$ \textcolor{orange!80!black}{\textsc{Rescue}}};

\node[io] (out) at (0,-10.0) {\textbf{Final verdict} $(\hat{y},\hat{w})$};

\draw[arr] (in.south) -| (u.north);
\draw[arr] (in.south) -| (wd.north);
\draw[arr] (u)  -- (sp);
\draw[arr] (sp) -- (mx);
\draw[arr] (mx) -- (rs);
\draw[arr] (rs) -- (wr);
\draw[arr] (wd) -- (sw);
\draw[arr] (sw) -- (pk);

\draw[arr] (wr.south) -- (-2.4,-6.6);
\draw[arr] (pk.south) |- (2.4,-6.6);
\draw[arr] (0,-9.0) -- (out.north);

\end{tikzpicture}
}
\caption{SIFT with claim-conditioned re-scoring. The structured stream decomposes the claim into active 5W1H units, extracts verbatim spans, and re-evaluates them against the \emph{full} claim to repair the verdict (orange). An independent NLI sweep over three-sentence windows feeds a late-stage verifier (deterministic Design~B or generative Design~A) that may keep, veto, or rescue the SIFT verdict. \textbf{Notation:} $c$ claim, $\mathcal{E}{=}\{e_i\}$ evidence pool, $u_j$ active 5W1H units, $S[i,j]/K[i,j]$ support/contradiction indicators, $s_{\max}/k_{\max}$ peak NLI entailment/contradiction over windows, $v_A$ SIFT-stage verdict, $\hat{y}/\hat{w}$ final verdict and warrant.}
\label{fig:method}
\vspace{-15pt}
\end{figure}
SIFT extracts verbatim 5W1H spans, applies claim-conditioned re-scoring to repair the verdict, and feeds the result to a deterministic late-stage verifier.

\subsection{SIFT: Structured Extraction is Not Enough}
Our choice of 5W1H  over adaptive atomic decomposition was guided by the explicit requirement of a typed, fixed-arity slot matrix that enables the typed failure taxonomy of Table \ref{tab:taxonomy}  (see Appendix \ref{app:atomic} for the comparison). SIFT first selects the active 5W1H units of the claim — inactive units are ignored rather than forced. For each active unit $u_j$ and each evidence item $e_i\in\mathcal{E}$, the backbone extracts a verbatim span when $e_i$ supports or contradicts $u_j$, or marks the unit unsupported. This produces a coverage matrix with support indicators $S[i,j]$ and contradiction indicators $K[i,j]$, exposing which facets are covered, by which documents, and whether the support is concentrated or patchwork. The matrix records coverage; coverage is not admissibility.

\subsection{Claim-Conditioned Re-scoring}

Once the 5W1H matrix has been aggregated into a base verdict, claim-conditioned re-scoring re-examines that verdict against the \emph{full} claim $c$: the backbone checks whether the evidence holds a grounded span that explicitly licenses or contradicts $c$ as a whole, not merely a parent facet $u_j$. The verdict is then revised to agree with that claim-level check---promoted when a grounded span licenses the full claim, demoted when one contradicts it. The warrant stays decomposed and traceable, but the final verdict has passed a claim-level compatibility test. This deliberate re-evaluation is what transforms decomposition from a liability—where isolated facets lose their context—into a reliable structural diagnostic. This step is part of SIFT itself, not of the downstream verifier layers (Design~A and Design~B); Algorithm~\ref{alg:sift} in Appendix~\ref{app:sift} gives the full procedure.

\subsection{Aggregation and the NLI Sweep}

The 5W1H matrix is aggregated into a base verdict: \textsc{Supports} when active units are covered by extracted spans and contradiction does not defeat the warrant, non-support otherwise. SIFT emits $\hat{w}$ as the connected premise built from retained spans. In parallel, we segment retrieved documents into three-sentence windows and score each window against the full claim with DeBERTa NLI \citep{laurer_less_2023}, retaining peak entailment $s_{\max}$ and peak contradiction $k_{\max}$. This sweep is not a substitute for SIFT---raw NLI applied to an open evidence pool cannot distinguish a coherent multi-axis warrant from a lexical match---but provides external semantic pressure on the repaired representation.

\subsection{Design B and Design A as Downstream Layers}
While SIFT repairs facet-level context, it does not inherently protect against patchwork aggregation across multiple documents. To bound this risk, we introduce a late-stage verifier layer.

\textbf{Design~B} applies a bounded rule over the SIFT verdict $v_A$, the SIFT warrant scores, and the independent sweep peaks. It computes the strongest entailment and contradiction evidence available to either stream, then chooses \textsc{Keep}, \textsc{Veto}, or \textsc{Rescue}. A veto can demote fragile \textsc{Supports} when sweep entailment is weak, or contradiction is strong, unless the warrant is protected by high entailment; a rescue can promote pre-declared missing-evidence or false-axis non-supports when entailment is high, and contradiction is low. Thresholds are selected on a 30\% validation split by pooling across primary backbones \emph{and} across datasets, yielding a single global tuple that is frozen for the held-out evaluation reported in the main tables. Algorithm~\ref{alg:designb} states the full rule.

\textbf{Design~A} is a generative comparison: an LLM adjudicator reads $c$, $v_A$, $\hat{w}$, the matrix, and the peak NLI scores, and outputs a decision via prompting. \emph{Rescue-only} can promote non-supports; \emph{two-sided} can additionally demote fragile \textsc{Supports}. The contrast we want is not ``LLM vs.\ rule''; it is whether the post-SIFT decision boundary is better expressed as a calibrated generative judgement or as a reproducible threshold rule.

\vspace{-10pt}
\section{Experimental Setup}
\label{sec:setup}
\vspace{-5pt}
\paragraph{Datasets.}
We examine four fact-checking datasets across two evidence regimes. \textbf{5PILS} \citep{tonglet2025cove} and \textbf{DP} \citep{crave_arka} are multi-evidence with retrieved pools and cross-document aggregation risk; \textbf{FEVER} \citep{thorne2018fever} and \textbf{SciFact} \citep{wadden2020scifact}  are single-evidence-oriented, where claims are usually licensed by one concise passage. For FEVER, we limit the evaluation to 1000 samples from the validation set.
\vspace{-5pt}
\paragraph{Backbones.}
Primary: \textbf{Qwen3-4B} \citep{qwen3technicalreport}, \textbf{Llama3.1-8B} \citep{grattafiori2024llama}, \textbf{Gemma2-9B} \citep{team2024gemma}; \textbf{Qwen3-14B} is a scaling backbone in the verifier table. We compare direct prompting, naive 5W1H, SIFT, Design A (rescue-only and two-sided), and Design B.
\vspace{-5pt}
\paragraph{Baselines.}
We compare \textbf{SIFT} against \textbf{Direct} prompting, which emits a verdict and a cited evidence span in a single pass, and \textbf{naive 5W1H}, which decomposes the claim and evidence without claim-conditioned re-scoring. The Direct baseline uses the model's cited span  as $\hat{w}$ and is scored on WSP only; the NLI sweep and Design~B are not applied.
\paragraph{Evaluation splits.}
The decomposition-stage comparison in Table~\ref{tab:decomp-repair} (Direct, naive 5W1H, SIFT) involves no tuned thresholds and is reported on the full evaluation set. The verifier table (Table~\ref{tab:verifier}) introduces Design~B, which uses thresholds selected on a 30\% validation split; to avoid leakage and give a fair comparison, all systems in that table---including the SIFT and Direct prompting baselines---are reported on the same 70\% held-out test split. SIFT and Direct prompting values, therefore, differ slightly between the two tables by split, not by computation.
\vspace{-5pt}
\paragraph{Metrics.}
We report binary accuracy and WSP. Paired accuracy comparisons use exact McNemar \citep{mcnemar1947note} with BH adjustment\citep {benjamini1995controlling} over the evaluated cells. We additionally validate WSP with (i) MiniCheck recomputation on every cell, (ii) a blind Gemini~2.5~Flash LLM-PPI(Prediction-Powered Inference) premise audit \citep{angelopoulos2024ppipp}, and (iii) a raw NLI-as-judge baseline.

\vspace{-5pt}
\section{Results}
\label{sec:results}

\vspace{-5pt}
\subsection{Re-scoring Repairs Rigid 5W1H Slotting}
\vspace{-5pt}

Naive 5W1H exposes warrant structure but fragments the verdict (Table~\ref{tab:decomp-repair}). The damage is largest on short-claim, single-evidence regimes, where the dataset does not redundantly cover multiple facets: FEVER loses up to 28 accuracy points under models such as Gemma2-9B. The cause is structural---locally scored facets become underspecified propositions, and aggregation propagates the local errors---not that decomposition is useless.

Claim-conditioned re-scoring restores the semantic context facet scoring removed without abandoning the 5W1H scaffold. Across the ten cells where naive 5W1H underperforms Direct, SIFT recovers $53.2\%$ of the lost accuracy (capped at the Direct baseline) and exceeds Direct in two. On the retrieved-evidence regime (5PILS, DP), naive 5W1H loses less ground to begin with, and SIFT's accuracy recovery is correspondingly smaller---it trails Direct in four of six cells, while SIFT WSP remains at or above Direct in five of six cells. The residual accuracy gap is not pure loss: in seven of the eight cells where SIFT trails Direct, Direct also carries the lower WSP, so part of its extra correct \textsc{Supports} are inadmissible true positives that the benchmark still credits. SIFT also misses some genuine supports; we therefore treat decomposition and re-scoring as a coupled method.
\begin{table}[t]
\centering
\resizebox{\columnwidth}{!}{%
\scriptsize
\setlength{\tabcolsep}{4.5pt}
\begin{tabular}{@{}llcccccccc@{}}
\toprule
 & & \multicolumn{2}{c}{\textbf{5PILS}} & \multicolumn{2}{c}{\textbf{DP}} & \multicolumn{2}{c}{\textbf{FEVER}} & \multicolumn{2}{c}{\textbf{SciFact}} \\
\cmidrule(lr){3-4}\cmidrule(lr){5-6}\cmidrule(lr){7-8}\cmidrule(lr){9-10}
\textbf{Model} & \textbf{Stage} & Acc & WSP & Acc & WSP & Acc & WSP & Acc & WSP \\
\midrule
\multirow{3}{*}{Qwen3-4B}
 & Direct                  & 71.8 & 50.2 & 57.9 & 45.9 & 88.2 & 68.7 & 69.7 & 48.1 \\
 & +\,naive 5W1H           & 71.1 & 54.4 & 41.5 & 29.7 & 64.6 & 56.1 & 71.3 & 43.9 \\
 \rowcolor{crowhi} & +\,claim re-score (\textbf{SIFT}) & \textbf{74.3} & \textbf{55.8} & \textbf{52.2} & \textbf{44.9} & \textbf{84.5} & \textbf{72.6} & \textbf{75.5} & \textbf{48.3} \\
\midrule
\multirow{3}{*}{Llama3.1-8B}
 & Direct                  & 76.4 & 34.9 & 67.4 & 29.7 & 81.7 & 47.8 & 67.6 & 29.9 \\
 & +\,naive 5W1H           & 69.3 & 66.4 & 51.4 & 72.5 & 69.0 & 59.5 & 56.4 & 55.6 \\
 \rowcolor{crowhi} & +\,claim re-score (\textbf{SIFT}) & \textbf{70.5} & \textbf{66.9} & \textbf{53.8} & \textbf{68.1} & \textbf{74.7} & \textbf{62.7} & \textbf{56.9} & \textbf{54.8} \\
\midrule
\multirow{3}{*}{Gemma2-9B}
 & Direct                  & 74.8 & 44.8 & 61.9 & 37.2 & 88.7 & 49.7 & 76.1 & 40.0 \\
 & +\,naive 5W1H           & 66.2 & 58.7 & 66.3 & 54.5 & 61.1 & 50.5 & 68.6 & 40.9 \\
 \rowcolor{crowhi} & +\,claim re-score (\textbf{SIFT}) & \textbf{69.9} & \textbf{56.3} & \textbf{67.4} & \textbf{49.8} & \textbf{78.7} & \textbf{59.3} & \textbf{77.7} & \textbf{44.6} \\
\bottomrule
\end{tabular}}
\caption{\textbf{Decomposition hurts; claim-conditioned re-scoring repairs.} Naive 5W1H drops Qwen3-4B/FEVER by 23.6 and Gemma2-9B/FEVER by 27.6 points. SIFT recovers a substantial share of the loss while keeping admissibility (WSP) gains. The shaded SIFT row is the repaired representation passed to downstream verifiers. Reported on the full evaluation set (no tuned thresholds in this comparison); for the same SIFT row on the 70\% held-out test split used by Design~B, see Table~\ref{tab:verifier}.}
\label{tab:decomp-repair}
\vspace{-15pt}
\end{table}

\begin{table}[t]
\centering
\scriptsize
\setlength{\tabcolsep}{4pt}
\begin{tabular}{@{}p{2.35cm}p{4.65cm}@{}}
\toprule
\textbf{Diagnostic claim} & \textbf{Evidence (source)} \\
\midrule
Rigid slotting hurts. & Naive 5W1H: Qwen3-4B/FEVER $-23.6$pt, Gemma2-9B/FEVER $-27.6$pt (Tab.~\ref{tab:decomp-repair}). \\
Repair needs the full claim. & SIFT recovers 53.2\% of the loss vs Direct (Tab.~\ref{tab:decomp-repair}). \\
Accuracy hides warrant quality. & Llama3.1-8B/DP Direct: 67.4 Acc, 29.7 WSP (Tab.~\ref{tab:decomp-repair}). \\
Verifier fixes combinations. & Design~B: WSP up 10/12 (Tab.~\ref{tab:verifier}); residuals shift away from aggregation (Fig.~\ref{fig:error-shift-models}). \\
WSP is scorer-robust. & MiniCheck: Design~B$>$SIFT in 13/16 cells (Tab.~\ref{tab:verifier}, MC columns); gold AUC $0.92$ (Tab.~\ref{tab:wsp-gold}). \\
\bottomrule
\end{tabular}
\caption{\textbf{Diagnostic summary (index).} The results are organised around five hypotheses about fallible warrant shortcuts, rather than around a single leaderboard. Each evidence cell points to the table or figure where the cited number appears.}
\label{tab:diagnostic-summary}
\vspace{-20pt}
\end{table}

\subsection{Verifier as a Downstream Operating Point}
Once the representation is repaired, Design B usually improves admissibility and often accuracy (Table~\ref{tab:verifier}). Across 12 primary held-out cells, Design~B improves accuracy in 9 and WSP in 10 relative to SIFT; paired exact McNemar tests give three BH-significant accuracy gains and no significant losses (Appendix~\ref{app:mcnemar}, Table~\ref{tab:mcnemar-app}). The one tie cell, Gemma2-9B/FEVER, is a passive safety-net case: FEVER produces no fragile SIFT-\textsc{Supports} and Gemma's rescue-eligible non-supports are NLI-confirmed, leaving the global tuple with nothing to act on  (Appendix~\ref{app:designb-actions}).
Against the Design~A two-sided adjudicator, Design~B is never strictly worse on WSP across all 16 held-out cells (12 strict wins, 4 ties).

\begin{table*}[t]
\centering
\resizebox{\textwidth}{!}{%
\begin{tabular}{@{}llrrrrrrrrrrrrrrrrrrrr@{}}
\toprule
 & & \multicolumn{5}{c}{\textbf{5PILS}} & \multicolumn{5}{c}{\textbf{DP}} & \multicolumn{5}{c}{\textbf{FEVER}} & \multicolumn{5}{c}{\textbf{SciFact}} \\
\cmidrule(lr){3-7}\cmidrule(lr){8-12}\cmidrule(lr){13-17}\cmidrule(lr){18-22}
\textbf{Model} & \textbf{System} & Acc & WSP & $\Delta$ & MC & $\Delta_{\text{M}}$ & Acc & WSP & $\Delta$ & MC & $\Delta_{\text{M}}$ & Acc & WSP & $\Delta$ & MC & $\Delta_{\text{M}}$ & Acc & WSP & $\Delta$ & MC & $\Delta_{\text{M}}$ \\
\midrule
\multirow{4}{*}{Qwen3-4B}
 & Direct              & 71.5 & 46.7 & & -- & & 56.5 & 43.2 & & -- & & 88.5 & 71.5 & & 61.1 & & 71.8 & 45.6 & & 50.9 & \\
 & SIFT                & 73.9 & 55.5 & & 46.0 & & 52.9 & 42.7 & & 38.0 & & 85.3 & 75.5 & & 62.5 & & 79.4 & 48.2 & & 47.1 & \\
 & Design A rescue     & \textbf{75.0} & 56.2 & {\small+0.7} & 46.1 & {\small+0.1}\,\cmark & \textbf{58.7} & 51.0 & {\small+8.3} & 43.2 & {\small+5.2}\,\cmark & 88.8 & 75.1 & {\small-0.4} & 61.5 & {\small-1.1}\,\cmark & 83.2 & 50.0 & {\small+1.8} & 48.9 & {\small+1.8}\,\cmark \\
 & Design A 2-sided    & 74.0 & 72.5 & {\small+17.0} & 60.1 & {\small+14.1}\,\cmark & 57.9 & 61.1 & {\small+18.4} & 52.4 & {\small+14.4}\,\cmark & 88.8 & 75.1 & {\small-0.4} & 61.5 & {\small-1.1}\,\cmark & 83.2 & 50.0 & {\small+1.8} & 48.9 & {\small+1.8}\,\cmark \\
 \rowcolor{crowhi} & Design B & 73.9 & \textbf{77.5} & {\small+22.0} & \textbf{61.5} & {\small+15.5}\,\cmark & 56.5 & \textbf{63.7} & {\small+21.0} & \textbf{52.2} & {\small+14.2}\,\cmark & \textbf{88.8} & 75.1 & {\small-0.4} & 61.5 & {\small-1.0}\,\cmark & \textbf{84.0} & \textbf{50.5} & {\small+2.3} & \textbf{48.4} & {\small+1.3}\,\cmark \\
\midrule
\multirow{4}{*}{Llama3.1-8B}
 & Direct              & 76.5 & 33.2 & & -- & & 67.5 & 28.9 & & -- & & 82.3 & 50.8 & & 45.5 & & 67.9 & 30.9 & & 27.9 & \\
 & SIFT                & 70.5 & 65.9 & & 57.4 & & 52.9 & 67.6 & & 59.9 & & 74.9 & 64.4 & & 53.0 & & 58.8 & 58.0 & & 58.0 & \\
 & Design A rescue     & 71.5 & 66.2 & {\small+0.3} & 57.4 & {\small\phantom{+}0.0}\,\cmark & 54.8 & 69.4 & {\small+1.8} & 60.1 & {\small+0.3}\,\cmark & 79.2 & 64.6 & {\small+0.2} & 53.0 & {\small\phantom{+}0.0}\,\cmark & 64.1 & 59.6 & {\small+1.6} & 58.0 & {\small\phantom{+}0.0}\,\cmark \\
 & Design A 2-sided    & 68.0 & 79.3 & {\small+13.4} & 57.5 & {\small+0.1}\,\cmark & 50.7 & 79.6 & {\small+12.0} & 60.1 & {\small+0.3}\,\cmark & 79.2 & 64.6 & {\small+0.2} & 53.0 & {\small\phantom{+}0.0}\,\cmark & 64.1 & 59.6 & {\small+1.6} & 58.0 & {\small\phantom{+}0.0}\,\cmark \\
 \rowcolor{crowhi} & Design B & \textbf{73.7} & \textbf{86.6} & {\small+20.7} & \textbf{70.6} & {\small+13.2}\,\cmark & \textbf{57.7} & \textbf{81.5} & {\small+13.9} & \textbf{71.2} & {\small+11.3}\,\cmark & \textbf{83.0}$^{\dagger}$ & \textbf{67.0} & {\small+2.6} & \textbf{54.6} & {\small+1.6}\,\cmark & \textbf{70.2}$^{\dagger}$ & \textbf{63.1} & {\small+5.1} & 56.9 & {\small-1.1}\,\xmark \\
\midrule
\multirow{4}{*}{Gemma2-9B}
 & Direct              & 75.3 & 43.9 & & -- & & 61.1 & 34.9 & & -- & & 88.1 & 49.6 & & 41.0 & & 77.9 & 38.7 & & 32.0 & \\
 & SIFT                & 70.4 & 55.6 & & 50.9 & & 67.5 & 48.8 & & 47.7 & & 78.2 & 60.2 & & 48.3 & & 78.6 & 46.1 & & 47.1 & \\
 & Design A rescue     & 71.0 & 56.0 & {\small+0.4} & 51.1 & {\small+0.2}\,\cmark & 67.8 & 48.6 & {\small-0.2} & 47.5 & {\small-0.1}\,\cmark & 78.2 & 60.2 & {\small\phantom{+}0.0} & 48.3 & {\small\phantom{+}0.0}\textsuperscript{tie} & \textbf{79.4} & \textbf{46.6} & {\small+0.5} & 47.6 & {\small+0.5}\,\cmark \\
 & Design A 2-sided    & \textbf{77.0} & 63.6 & {\small+8.0} & 53.7 & {\small+2.8}\,\cmark & \textbf{68.5} & 52.7 & {\small+3.9} & 50.8 & {\small+3.1}\,\cmark & 78.2 & 60.2 & {\small\phantom{+}0.0} & 48.3 & {\small\phantom{+}0.0}\textsuperscript{tie} & \textbf{79.4} & \textbf{46.6} & {\small+0.5} & 47.6 & {\small+0.5}\,\cmark \\
 \rowcolor{crowhi} & Design B & 72.8 & \textbf{71.4} & {\small+15.8} & \textbf{60.7} & {\small+9.8}\,\cmark & 63.7 & \textbf{68.0} & {\small+19.2} & \textbf{64.5} & {\small+16.8}\,\cmark & 78.2 & 60.2 & {\small\phantom{+}0.0} & 48.3 & {\small\phantom{+}0.0}\textsuperscript{tie} & \textbf{79.4} & \textbf{46.6} & {\small+0.5} & \textbf{47.6} & {\small+0.5}\,\cmark \\
\midrule
\multirow{4}{*}{Qwen3-14B}
 & Direct              & 70.7 & 46.0 & & -- & & 56.0 & 34.2 & & -- & & 95.1 & 55.3 & & -- & & 79.4 & 41.5 & & -- & \\
 & SIFT                & 79.2 & 63.8 & & 56.8 & & 67.5 & 60.3 & & 56.8 & & 94.0 & 79.1 & & 63.0 & & 80.9 & 64.6 & & 63.1 & \\
 & Design A rescue     & \textbf{79.6} & 64.4 & {\small+0.6} & 56.9 & {\small+0.1}\,\cmark & \textbf{69.0} & 60.6 & {\small+1.5} & 56.9 & {\small+0.1}\,\cmark & 94.4 & 79.3 & {\small+0.2} & 63.3 & {\small+0.3}\,\cmark & 84.0 & 66.7 & {\small+2.1} & 63.8 & {\small+0.7}\,\cmark \\
 & Design A 2-sided    & 77.4 & 65.4 & {\small+1.6} & 56.2 & {\small-0.7}\,\xmark & 66.6 & 61.6 & {\small+1.3} & 56.8 & {\small\phantom{+}0.0}\,\cmark & 94.4 & 79.3 & {\small+0.2} & 63.3 & {\small+0.3}\,\cmark & 84.0 & 66.7 & {\small+2.1} & 63.8 & {\small+0.7}\,\cmark \\
 \rowcolor{crowhi} & Design B & 77.4 & \textbf{69.6} & {\small+5.8} & \textbf{58.5} & {\small+1.7}\,\cmark & 66.8 & \textbf{62.1} & {\small+1.8} & \textbf{57.3} & {\small+0.5}\,\cmark & \textbf{94.7} & \textbf{79.6} & {\small+0.5} & \textbf{63.4} & {\small+0.4}\,\cmark & 84.0 & 66.7 & {\small+2.1} & \textbf{63.8} & {\small+0.7}\,\cmark \\
\midrule
\multicolumn{2}{@{}l}{NLI-as-judge (DeBERTa)} & 63.9 & -- & -- & -- & -- & 53.6 & -- & -- & -- & -- & 91.7 & -- & -- & -- & -- & 69.7 & -- & -- & -- & -- \\
\bottomrule
\end{tabular}%
}
\caption{\textbf{Verifier over repaired SIFT, held-out test.} $\Delta$ is the primary WSP change relative to SIFT in the same row block (default: DeBERTa). To verify scorer robustness, MC displays the absolute WSP scored by the independent grounding model MiniCheck, alongside its delta ($\Delta_{\text{M}}$). The \cmark/\xmark\ symbols indicate sign agreement between the DeBERTa and MiniCheck deltas across Design~A and Design~B rows (agrees in 43 of 45 cells where both deltas are non-zero; 3 tied cells with $\Delta=\Delta_{\text{M}}=0$). Design~B improves WSP in 10 of 12 primary cells and accuracy in 9 of 12; ${}^{\dagger}$marks BH-significant accuracy gains over SIFT (Appendix~\ref{app:mcnemar}). The Gemma2-9B/DP accuracy drop is not significant. Qwen3-14B is a scaling backbone. The NLI-as-judge baseline applies DeBERTa directly without SIFT; WSP is omitted because its cited warrant is the NLI-selected span itself.}
\vspace{-15pt}
\label{tab:verifier}
\end{table*}

\vspace{-10pt}
\paragraph{The accuracy--admissibility trade-off.}
Against direct prompting, Design~B trades accuracy for admissibility on five cells (Table~\ref{tab:verifier}), giving up 9--11 accuracy points on the two largest trades (Llama3.1-8B/DP, Gemma2-9B/FEVER) while raising WSP by 11--52. All five trade cells share one signature: Direct's WSP is below $50$ while its accuracy is competitive. The accuracy Design B forfeits therefore comes mostly from the low-WSP layer where Direct relies on inadmissible Supports—binary accuracy counts every correct SUPPORTS equally, including labels whose cited warrant does not license the claim. The trade is not universally benign: over-demoting borderline supports also creates genuine false negatives (Gemma2-9B/DP). An accuracy delta is uninterpretable on its own; paired with WSP, it separates correcting inadmissible true positives from losing recall. On the Qwen3-14B backbone, Design~B's accuracy deltas flatten while WSP still wins 4 of 4---as base models improve, the marginal accuracy benefit of deterministic stabilisation narrows while admissibility remains useful for auditing.
\vspace{-10pt}
\paragraph{Generative adjudication is competitive but less stable.}
Rescue-only Design~A improves accuracy over SIFT in 11 of 12 primary cells with modest WSP movement; two-sided has a higher WSP ceiling but the demotion path is model- and regime-sensitive (see Limitations, \S\ref{lim:designa}), so Design~B supplies the bounded operating point.

\subsection{WSP Survives a Change of NLI Family}
\label{sec:wsp-valid}
The systemic gains above rely on WSP accurately reflecting admissibility. To confirm WSP is a robust diagnostic, we first test whether its rankings survive a change of NLI family. We recompute every SIFT and Design~B WSP value with MiniCheck-Flan-T5-Large \citep{tang2024minicheck}, a separately trained grounding classifier. The ranking survives: global Spearman correlation \citep{spearman1904proof} with DeBERTa-WSP is $\rho=0.89$ over 32 cells, and MiniCheck preserves the Design~B-over-SIFT direction in 13 of 16 held-out cells (claim summarised in Table~\ref{tab:diagnostic-summary}; per-cell numbers in Table~\ref{tab:minicheck-app}); the largest MiniCheck lifts also track the largest DeBERTa lifts. Because DeBERTa is the feature provider \emph{inside} Design~B, this cross-family agreement isolates the WSP improvement as a property of the repaired representation, not of reusing one NLI head as feature and scorer.

\begin{table}[t]
\centering
\scriptsize
\setlength{\tabcolsep}{4pt}
\begin{tabular}{@{} p{2.6cm} p{2.2cm} p{2.5cm} @{}}
\toprule
\textbf{Method \& Scope} & \textbf{Estimate / Metric} & \textbf{Key Finding} \\
\midrule
\multicolumn{3}{@{}l}{\textit{External premise audits}} \\
\textbf{Gemini LLM-PPI++} \newline \textit{\scriptsize 100 Qwen3-4B items} & 58.2 [51.3, 65.1] \newline $\phi=0.720$ & 86\% agreement.  \\
\midrule
\multicolumn{3}{@{}l}{\textit{Scorer robustness}} \\
\textbf{MiniCheck} \newline \textit{\scriptsize 32 SIFT/B cells} & global $\rho=0.89$ \newline vs DeBERTa-WSP & Gains persist: B $>$ SIFT in 13/16 cells. \\
\midrule
\multicolumn{3}{@{}l}{\textit{Dataset LLM-PPI++ estimates (random subsets)}} \\
\textbf{5PILS} & 48.6 [36.9, 60.2] & Multi-evidence claims. \\
\textbf{DP} & 45.0 [29.3, 60.7] & Temporal/event claims. \\
\textbf{FEVER} & 77.2 [64.5, 89.9] & Near NLI train dist. \\
\textbf{SciFact} & 48.3 [34.4, 62.2] & Scientific warrants. \\
\bottomrule
\end{tabular}
\caption{\textbf{Validation checks for WSP.} External audits estimate whether the exact warrant premise licenses the claim, while scorer-robustness tests whether WSP trends survive an independent grounding model. Dataset-level estimates are shown separately; agreement with WSP is reported only at the pooled level because per-dataset subsets are small.}
\label{tab:wsp-validation}
\vspace{-20pt}
\end{table}

\paragraph{External audits.}
Beyond the MiniCheck NLI-family check above, a blind Gemini~2.5~Flash LLM-PPI++ audit on 100 stratified random Qwen3-4B premises estimates pooled admissibility at $58.2$ [51.3, 65.1], with Pearson's phi coefficient $\phi=0.720$ \citep{pearson1900correlation} and 86\% agreement against WSP labels. The Gemini level sits below some DeBERTa-WSP values because the two scorers set the binary decision boundary differently at the margin; their relative rankings 
agree ($\phi=0.720$, 86\% agreement), and absolute calibration against human gold evidence is established separately  (§\ref{sec:wsp-gold}, Table~\ref{tab:wsp-gold}). The per-dataset Gemini estimates on retrieved evidence (5PILS $48.6$, DP  $45.0$) remain within $\sim$10 absolute points of the corresponding DeBERTa-WSP with overlapping CIs; we therefore use WSP as a \emph{relative} diagnostic for comparing systems under the same scorer, not as an absolute human-admissibility measure. Per-dataset estimates are tabulated in Table~\ref{tab:wsp-validation}.

\subsection{WSP Agrees with Human Gold Evidence}
\label{sec:wsp-gold}
To move beyond the above model-mediated checks, we calibrate WSP against the human-annotated gold evidence and verdict labels native to FEVER and SciFact. In our pipeline, each record for these datasets contains exactly one evidence item, verified to be the gold sentence verbatim. The scored warrants are therefore human-curated text, and a predicted-\textsc{Supports} verdict is human-admissible exactly when the gold label is \textsc{Supports}. Table~\ref{tab:wsp-gold} stratifies every predicted-\textsc{Supports} verdict on FEVER and SciFact by that label (SIFT stage, three LLM backbones). WSP's entailment score separates human-admissible from inadmissible supports with AUC $0.92$ under DeBERTa and $0.91$ under MiniCheck; when WSP calls a support \emph{warranted}, the claim is human-admissible $97.8\%$ of the time, well above the $76\%$ base rate, and the two independently trained NLI families agree on this calibration. WSP also errs toward caution: it confirms $79\%$ of genuinely admissible supports but fires on only $5.6\%$ of inadmissible ones, so a reported WSP value is a conservative lower bound on admissibility, not an inflated one.  Two conditions bound this result. It covers the two datasets with sentence-level human rationales; 5PILS and DP use retrieved evidence and admit no comparable reference. And because the gold warrant is a single well-formed sentence, the comparison tests whether WSP agrees with human judgement on clean evidence rather than whether it closes the warrant gap under noisy retrieval. Within those bounds, WSP's \textsc{Supports}-licensing verdicts align with human gold annotation, and the alignment is slightly stronger on SIFT over Direct prompting (pooled AUC $0.92$ vs. $0.83$)
(Table~\ref{tab:wsp-gold}).

\begin{table}[t]
\centering
\scriptsize
\setlength{\tabcolsep}{5pt}
\resizebox{0.7\columnwidth}{!}{%
\begin{tabular}{@{} l r c c c c @{}}
\toprule
\textbf{Subset} & $n$ & \textbf{AUC} & \textbf{Prec.} & \textbf{WSP$\mid$g$^{+}$} & \textbf{WSP$\mid$g$^{-}$} \\
\midrule
\multicolumn{6}{@{}l}{\textit{DeBERTa-WSP (Direct stage)}} \\
\quad FEVER   & 1388 & 0.838 & 0.983 & 0.616 & 0.081 \\
\quad SciFact &  279 & 0.775 & 0.972 & 0.430 & 0.086 \\
\quad Pooled  & 1667 & 0.826 & 0.982 & 0.585 & 0.082 \\
\addlinespace
\multicolumn{6}{@{}l}{\textit{DeBERTa-WSP (SIFT stage)}} \\
\quad FEVER   & 1695 & 0.937 & 0.977 & 0.832 & 0.060 \\
\quad SciFact &  341 & 0.896 & 0.988 & 0.596 & 0.029 \\
\quad Pooled  & 2036 & 0.921 & 0.978 & 0.790 & 0.056 \\
\addlinespace
\multicolumn{6}{@{}l}{\textit{MiniCheck-WSP (Direct stage)}} \\
\quad FEVER   & 1388 & 0.866 & 0.984 & 0.534 & 0.069 \\
\quad SciFact &  279 & 0.778 & 0.948 & 0.377 & 0.143 \\
\quad Pooled  & 1667 & 0.849 & 0.979 & 0.508 & 0.082 \\
\addlinespace
\multicolumn{6}{@{}l}{\textit{MiniCheck-WSP (SIFT stage)}} \\
\quad FEVER   & 1695 & 0.922 & 0.984 & 0.689 & 0.034 \\
\quad SciFact &  341 & 0.812 & 0.931 & 0.548 & 0.159 \\
\quad Pooled  & 2036 & 0.908 & 0.976 & 0.664 & 0.051 \\
\addlinespace
\bottomrule
\end{tabular}}
\caption{\textbf{WSP calibrated against human gold evidence.} Each predicted-\textsc{Supports} verdict on FEVER/SciFact (three backbones pooled: Qwen3-4B, Llama3.1-8B, Gemma2-9B) is labelled human-admissible iff its gold verdict is \textsc{Supports}; the scored warrant is the dataset's gold evidence sentence. AUC is for WSP's entailment score against that label. \textbf{Prec.}\ is the fraction of WSP-\emph{warranted} supports that are gold-admissible (pooled base rate: $0.76$ at SIFT stage, $0.88$ at Direct stage; see Appendix~\ref{app:base-rates} for the definition and computation). WSP$\mid$g$^{+}$ and WSP$\mid$g$^{-}$ are WSP entailment rates on gold-admissible and gold-inadmissible supports. Direct stage $n$ is smaller than SIFT-stage $n$ because Direct prompting issues fewer predicted-\textsc{Supports} verdicts overall, and for Llama3.1-8B, we drop verdicts lacking an extractable cited span.}
\label{tab:wsp-gold}
\vspace{-15pt}
\end{table}
\vspace{-5pt}

\subsection{Residual Errors}
\begin{figure}[t]
\centering
\resizebox{0.8\columnwidth}{!}{%
\begin{tikzpicture}[
  x=0.72cm, y=0.085cm,  
  font=\footnotesize,   
  axis/.style={draw=black!50, semithick},
  tick/.style={draw=black!50},
  grid/.style={draw=black!8, thin},
  dslabel/.style={font=\normalsize\bfseries, align=center},
]

\def\bw{0.38}
\newcommand{\stackerr}[7]{%
  \pgfmathsetmacro{\yA}{#2}
  \pgfmathsetmacro{\yB}{#2+#3}
  \pgfmathsetmacro{\yC}{#2+#3+#4}
  \pgfmathsetmacro{\yD}{#2+#3+#4+#5}
  \pgfmathsetmacro{\yE}{#2+#3+#4+#5+#6}
  \filldraw[fill=cident,  draw=white, line width=0.4pt] (#1,0)   rectangle (#1+\bw,\yA);
  \filldraw[fill=cpred,   draw=white, line width=0.4pt] (#1,\yA) rectangle (#1+\bw,\yB);
  \filldraw[fill=caggr,   draw=white, line width=0.4pt] (#1,\yB) rectangle (#1+\bw,\yC);
  \filldraw[fill=cunder,  draw=white, line width=0.4pt] (#1,\yC) rectangle (#1+\bw,\yD);
  \filldraw[fill=cunable, draw=white, line width=0.4pt] (#1,\yD) rectangle (#1+\bw,\yE);
  \draw[black!25, line width=0.3pt] (#1,0) rectangle (#1+\bw,\yE);
  \node[rotate=45, anchor=east, font=\scriptsize, inner sep=1pt] at (#1+\bw/2, -1.5) {#7};
}

\begin{scope}[yshift=0cm]
  \draw[axis] (0,0) -- (16.2,0);
  \draw[axis] (0,0) -- (0,56);
  \foreach \v in {0,10,20,30,40,50} {
    \draw[tick] (-.15,\v) -- (.15,\v);
    \node[anchor=east, font=\footnotesize] at (-.35,\v) {\v};
    \draw[grid] (0,\v) -- (16.2,\v);
  }
  \node[rotate=90, anchor=south, font=\footnotesize] at (-1.8, 25) {Residual error rate (\%)};

  \stackerr{0.50}{1.60}{3.60}{11.00}{6.10}{3.40}{S}
  \stackerr{0.92}{0.40}{7.20}{6.30}{5.10}{2.80}{B}
  \stackerr{2.20}{4.30}{13.10}{9.30}{2.30}{0.50}{S}
  \stackerr{2.62}{2.20}{14.80}{4.80}{1.40}{0.30}{B}
  \stackerr{3.90}{4.00}{7.50}{17.20}{0.60}{0.80}{S}
  \stackerr{4.32}{2.30}{11.00}{8.10}{0.50}{0.70}{B}
  \stackerr{5.60}{0.60}{14.30}{2.60}{1.30}{1.00}{S}
  \stackerr{6.02}{0.50}{14.10}{2.50}{1.00}{0.70}{B}
  \node[dslabel] at (3.43, 46) {5PILS};
  \foreach \x/\m in {0.69/Q4, 2.39/L8, 4.09/G9, 5.79/Q14} {\node[font=\scriptsize, text=black!70] at (\x, -4.5) {\m};}

  \draw[black!12, dashed, thin] (8.2, -4) -- (8.2, 53); 

  \stackerr{9.00}{0.49}{6.65}{7.46}{31.28}{1.94}{S}
  \stackerr{9.42}{0.32}{6.97}{7.78}{22.85}{1.78}{B}
  \stackerr{10.70}{1.78}{25.45}{7.13}{11.51}{0.32}{S}
  \stackerr{11.12}{1.62}{24.47}{8.43}{8.27}{0.32}{B}
  \stackerr{12.40}{1.30}{6.32}{23.18}{0.65}{1.13}{S}
  \stackerr{12.82}{1.13}{14.26}{16.21}{0.49}{0.81}{B}
  \stackerr{14.10}{1.62}{23.50}{3.08}{2.92}{0.49}{S}
  \stackerr{14.52}{1.62}{21.39}{4.70}{2.11}{0.49}{B}
  \node[dslabel] at (11.93, 46) {DP};
  \foreach \x/\m in {9.19/Q4, 10.89/L8, 12.59/G9, 14.29/Q14} {\node[font=\scriptsize, text=black!70] at (\x, -4.5) {\m};}
\end{scope}

\begin{scope}[yshift=-6.8cm]
  \draw[axis] (0,0) -- (16.2,0);
  \draw[axis] (0,0) -- (0,56);
  \foreach \v in {0,10,20,30,40,50} {
    \draw[tick] (-.15,\v) -- (.15,\v);
    \node[anchor=east, font=\footnotesize] at (-.35,\v) {\v};
    \draw[grid] (0,\v) -- (16.2,\v);
  }
  \node[rotate=90, anchor=south, font=\footnotesize] at (-1.8, 25) {Residual error rate (\%)};

  \stackerr{0.50}{0.10}{2.30}{7.50}{5.10}{0.50}{S}
  \stackerr{0.92}{0.10}{2.40}{3.10}{1.70}{0.40}{B}
  \stackerr{2.20}{0.00}{1.30}{13.80}{9.30}{0.90}{S}
  \stackerr{2.62}{0.00}{1.10}{3.50}{1.90}{0.20}{B}
  \stackerr{3.90}{0.00}{0.80}{19.50}{0.60}{0.40}{S}
  \stackerr{4.32}{0.00}{0.70}{5.70}{0.60}{0.30}{B}
  \stackerr{5.60}{0.40}{2.10}{1.90}{1.50}{0.60}{S}
  \stackerr{6.02}{0.40}{1.50}{1.60}{1.10}{0.30}{B}
  \node[dslabel] at (3.43, 46) {FEVER};
  \foreach \x/\m in {0.69/Q4, 2.39/L8, 4.09/G9, 5.79/Q14} {\node[font=\scriptsize, text=black!70] at (\x, -4.5) {\m};}

  \draw[black!12, dashed, thin] (8.2, -4) -- (8.2, 53); 

  \stackerr{9.00}{0.00}{4.26}{11.17}{6.38}{2.66}{S}
  \stackerr{9.42}{0.00}{2.66}{5.85}{3.72}{2.66}{B}
  \stackerr{10.70}{0.00}{6.38}{7.98}{28.19}{0.53}{S}
  \stackerr{11.12}{0.00}{5.85}{2.13}{17.55}{0.53}{B}
  \stackerr{12.40}{0.00}{2.66}{17.02}{0.00}{2.66}{S}
  \stackerr{12.82}{0.00}{2.66}{6.38}{0.00}{2.13}{B}
  \stackerr{14.10}{0.00}{6.91}{2.13}{5.85}{4.79}{S}
  \stackerr{14.52}{0.00}{5.32}{1.06}{5.32}{4.26}{B}
  \node[dslabel] at (11.93, 46) {SciFact};
  \foreach \x/\m in {9.19/Q4, 10.89/L8, 12.59/G9, 14.29/Q14} {\node[font=\scriptsize, text=black!70] at (\x, -4.5) {\m};}
\end{scope}

\end{tikzpicture}
}
\caption{\textbf{Residual error shift by model.}
Each dataset contains four model columns (Q4=Qwen3-4B, L8=Llama3.1-8B, G9=Gemma2-9B, Q14=Qwen3-14B).
Within each column, \textbf{S} (left) is the SIFT baseline and
\textbf{B} (right) is Design~B.
Segments encode the five residual error types (bottom to top):
\textcolor{cident}{$\blacksquare$}~\textsc{Identity}, \textcolor{cpred}{$\blacksquare$}~\textsc{Predicate}, \textcolor{caggr}{$\blacksquare$}~\textsc{Aggregation},
\textcolor{cunder}{$\blacksquare$}~\textsc{Undercutter}, and \textcolor{cunable}{$\blacksquare$}~\textsc{Unable}.
Aggregation reductions are strongest for the smaller backbones
and attenuate for Qwen3-14B; predicate and identity errors
remain the main extraction-bound residual, with undercutter
errors forming a separate retrieval frontier.}
\label{fig:error-shift-models}
\vspace{-15pt}
\end{figure}
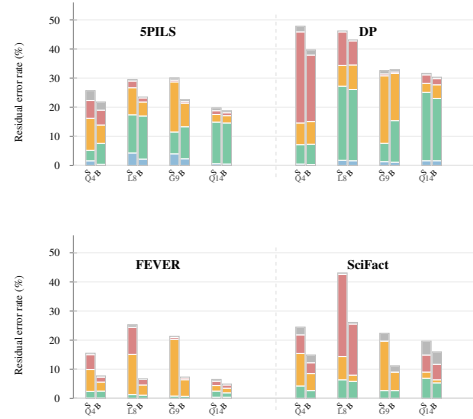

\vspace{-5pt}
Aggregation errors fall on every dataset; identity and predicate errors persist (Fig.~\ref{fig:error-shift-models}). The asymmetry has a structural cause: a verifier can keep, demote, or rescue, but cannot insert a span extraction that was never selected. \textsc{Identity} and \textsc{Predicate} errors bind the warrant to the wrong entity or property, so the licensing span is absent from the verifier's input. \textsc{Aggregation} errors are errors of \emph{combination} over correctly extracted spans, so a verifier that rejects incoherent combinations repairs them. This bounds what any post-hoc verifier---deterministic or generative---can achieve over a fixed extractor. \textsc{Undercutter} errors require omitted context that may not be in the cited warrant; they are a separate frontier.

\vspace{-10pt}
\section{Discussion}
\vspace{-10pt}
Decomposition is conditional: it exposes warrant structure but lowers accuracy when local facets are treated as complete semantic units (Tables~\ref{tab:decomp-repair} and \ref{tab:verifier}). The accuracy gains direct prompting reports can rest on inadmissible warrants, and WSP makes that gap measurable. Rigid 5W1H slotting then erases those same gains unless claim-conditioned re-scoring is folded back in. 
To rule out that decomposition itself is the culprit, we swap the fixed schema for atomic-claim decomposition \citep{min2023factscore,kamoi2023wice}, which recovers most of the gap on short claims (Table~\ref{tab:decomp-scheme}) --- the slot frame fails, not decomposition. We keep 5W1H because Design~B's rescue and veto rules act on slot identity, which atomic claims do not expose; the verifier itself can only stabilise the verdict; it cannot construct entities or predicates the extractor missed, so further gains lie in better warrant construction, not better adjudication.

\begin{table}[t]
\centering
\resizebox{\columnwidth}{!}{%
\scriptsize
\setlength{\tabcolsep}{5pt}
\begin{tabular}{@{}llcccc@{}}
\toprule
\textbf{Model} & \textbf{Decomposition} & \textbf{5PILS} & \textbf{DP} & \textbf{FEVER} & \textbf{SciFact} \\
\midrule
\multirow{3}{*}{Qwen3-4B}
 & none (Direct) & 71.8 & 57.9 & 88.2 & 69.7 \\
 & naive 5W1H    & 71.1 & \textbf{41.5} & \textbf{64.6} & 71.3 \\
 & atomic        & 66.5 & 57.4 & 87.2 & 72.3 \\
\midrule
\multirow{3}{*}{Gemma2-9B}
 & none (Direct) & 74.8 & 61.9 & 88.7 & 76.1 \\
 & naive 5W1H    & \textbf{66.2} & 66.3 & \textbf{61.1} & \textbf{68.6} \\
 & atomic        & 76.5 & 64.2 & 93.3 & 73.9 \\
\bottomrule
\end{tabular}}
\caption{\textbf{Rigid vs.\ adaptive decomposition (accuracy).} Naive 5W1H loses accuracy on short-claim regimes (bold), while atomic-claim decomposition---granularity-adaptive, so it does not over-slot simple claims---stays much closer to direct prompting. The fragmentation dilemma is a property of rigid slotting, not of decomposition itself; the full grid with WSP and claim-conditioned re-scoring is in Appendix~\ref{app:atomic}.}
\label{tab:decomp-scheme}
\vspace{-15pt}
\end{table}
\vspace{-5pt}

\section{Conclusion}
\vspace{-5pt}
Verdict accuracy is a sufficient measure of fact-checking only when the cited warrant licenses the claim, and on current benchmarks, it often does not. We supplied two pieces that turn this slack into something a system can act on: \textbf{SIFT},  which asks whether a structured rationale survives recombination with the full claim, and \textbf{WSP}, an automatic admissibility check that holds against human gold evidence on FEVER and SciFact. Together, they provide a paired axis on which structured decomposition stops looking like a trade-off between accuracy and admissibility and begins to look like a discipline that can be reported. The remaining headroom is upstream of the verifier: no post-hoc rule can reconstruct an entity or predicate that extraction never surfaced. Benchmarks that score only verdicts will continue to reward systems that reach the right label through inadmissible reasoning; reporting both axes is the change that makes the next round of progress legible.

\section*{Limitations}

\paragraph{WSP is a proxy.} WSP is an automatic diagnostic, not a bespoke expert admissibility audit. WSP is scored with DeBERTa NLI, and its signal is corroborated from three independent directions: a separately trained NLI family (MiniCheck), a blind Gemini LLM-PPI++ audit \citep{angelopoulos2024ppipp}, and direct calibration against the datasets' human gold annotation on FEVER and SciFact (\S\ref{sec:wsp-gold}), where WSP attains AUC $0.92$ and precision $0.98$ on $n=2036$ predicted-\textsc{Supports} verdicts. The first two are model-mediated; the third is human-curated but covers only the oracle-evidence datasets. We deliberately did not crowd-source admissibility for the retrieved-evidence datasets (5PILS, DP): identifying warrant-gap shortcuts, such as patchwork aggregation or subtle predicate mismatches across multiple retrieved documents, is an expert analytical task, and non-expert crowd labels in this format inject more noise than signal. A focused expert admissibility audit on those datasets, where no sentence-level human rationale exists, remains the appropriate next step.

\paragraph{NLI as feature and scorer.} Design~B uses DeBERTa NLI as a feature, and WSP uses NLI entailment as its scorer, so we decouple the two with cross-family checks. MiniCheck---a separately trained grounding model---preserves the Design~B-over-SIFT WSP direction in 13 of 16 cells, and a raw NLI-as-judge baseline shows that NLI without SIFT's facet matrix does not reproduce the structured gains. Because both models share an entailment-task lineage, expert annotation would further isolate task-family effects; the cross-family agreement suggests that the gains are not an artefact of a single NLI head.

\paragraph{Threshold calibration.} Design~B exposes a calibrated operating point: its thresholds are selected on a 30\% validation split to set the accuracy--admissibility balance, and a single global tuple stays competitive across datasets (Appendix~\ref{app:thresholds}). Deployments should calibrate on local validation data.

\paragraph{Generative adjudication: demotion-path sensitivity.}
\label{lim:designa}
The two-sided variant of Design~A reaches a higher WSP ceiling than Design~B when demotion calibrates---it helps Gemma2-9B/5PILS where Design~B loses 4.2 accuracy points to it under BH-adjusted McNemar---but the demotion path is model- and regime-sensitive: on Llama3.1-8B/5PILS, two-sided demotes 226 \textsc{Supports} that rescue-only keeps, producing a large net accuracy loss. 
\paragraph{Deployment considerations.} A practical playbook for adopting the stack: use SIFT when warrant traceability matters, irrespective of verifier choice; add Design~B when the target is a reproducible WSP operating point with bounded accuracy cost (calibrate on local validation data, Appendix~\ref{app:thresholds}); reserve Design~A two-sided for settings where the demotion path has been calibrated on matched validation data (\S\ref{lim:designa}). The verifier question is downstream of the representation question.

\paragraph{Scope of 5W1H.} The 5W1H schema is event-centric: its units fit claims about events, people, places, times, causes, and mechanisms---the compound, multi-axis claims SIFT targets, and the regime where the warrant gap is most consequential in real-time fact-checking. It is a less natural fit for abstract scientific relations, long causal chains, or claims whose warrant is statistical, and claim-conditioned re-scoring does not guarantee the chosen facets are the right semantic units for those. We include SciFact \citep{wadden2020scifact} as a boundary case: SIFT still applies, with smaller WSP gains than on event-centric data. The claim-conditioned re-scoring step itself transfers to adaptive (atomic) decomposition (Appendix~\ref{app:atomic}).

\paragraph{Decomposition-scheme choice.} The 5W1H schema is adopted not for raw accuracy---adaptive atomic decomposition can score higher on short-claim regimes---but because its typed, fixed-arity matrix is the prerequisite for deterministic verification and typed error analysis, the substrate the rest of the pipeline requires. Atomic decomposition is granularity-adaptive: near-atomic FEVER claims stay close to a single fact while compound 5PILS claims split into roughly three, so it does not force slot structure onto claims that under-specify it. A complementary line learns the optimal atomicity per verifier via reinforcement learning \citep{lu2025optimizing}; we trade that flexibility for a deterministic, scaffold-preserving fix.

\paragraph{Extraction-bound residual errors.} Because the verifier neutralises aggregation errors, the residual errors are extraction-bound---identity and predicate errors: a post-hoc rule cannot reconstruct an entity or property the extractor never surfaced. Rather than a dead end, this cleanly isolates warrant construction and omission-aware retrieval (for undercutters) as the next research frontier.

\paragraph{Backbone and language scope.} Backbones differ in calibration, span-copying behaviour, and sensitivity to output contracts, even when evaluated under the same held-out protocol. All experiments are on English text. Cross-lingual evidence, multimodal claims, and languages with different argument structures may require different decomposition units and warrant displays.
\paragraph{AI assistance.} AI assistants were used in a limited editorial capacity --- sentence-level prose editing and rewriting only.

\bibliography{references}

\begin{thebibliography}{38}
\providecommand{\natexlab}[1]{#1}

\bibitem[{Abdelnabi et~al.(2022)Abdelnabi, Hasan, and Fritz}]{abdelnabi2022open}
Sahar Abdelnabi, Rakibul Hasan, and Mario Fritz. 2022.
\newblock Open-domain, content-based, multi-modal fact-checking of out-of-context images via online resources.
\newblock In \emph{Proceedings of the IEEE/CVF conference on computer vision and pattern recognition}, pages 14940--14949.

\bibitem[{Akhtar et~al.(2026)Akhtar, Schlichtkrull, and Vlachos}]{akhtar2026ev2r}
Mubashara Akhtar, Michael Schlichtkrull, and Andreas Vlachos. 2026.
\newblock \href {https://doi.org/10.1162/TACL.a.647} {Ev2r: Evaluating evidence retrieval in automated fact-checking}.
\newblock \emph{Transactions of the Association for Computational Linguistics}, 14:530--561.

\bibitem[{Angelopoulos et~al.(2023)Angelopoulos, Duchi, and Zrnic}]{angelopoulos2024ppipp}
Anastasios~N Angelopoulos, John~C Duchi, and Tijana Zrnic. 2023.
\newblock Ppi++: Efficient prediction-powered inference.

\bibitem[{Benjamini and Hochberg(1995)}]{benjamini1995controlling}
Yoav Benjamini and Yosef Hochberg. 1995.
\newblock Controlling the false discovery rate: a practical and powerful approach to multiple testing.
\newblock \emph{Journal of the Royal statistical society: series B (Methodological)}, 57(1):289--300.

\bibitem[{Choi et~al.(2021)Choi, Palomaki, Lamm, Kwiatkowski, Das, and Collins}]{choi2021decontextualization}
Eunsol Choi, Jennimaria Palomaki, Matthew Lamm, Tom Kwiatkowski, Dipanjan Das, and Michael Collins. 2021.
\newblock \href {https://doi.org/10.1162/tacl_a_00377} {Decontextualization: Making sentences stand-alone}.
\newblock \emph{Transactions of the Association for Computational Linguistics}, 9:447--461.

\bibitem[{Dey et~al.(2026)Dey, Awan, Channing, Witt, and Collomosse}]{crave_arka}
Arka~Ujjal Dey, Muhammad~Junaid Awan, Georgia Channing, Christian Schroeder~de Witt, and John Collomosse. 2026.
\newblock \href {https://doi.org/10.1109/TCSS.2026.3669799} {Fact-checking with contextual narratives: Leveraging retrieval-augmented llms for social media analysis}.
\newblock \emph{IEEE Transactions on Computational Social Systems}, pages 1--12.

\bibitem[{Gao et~al.(2023)Gao, Dai, Pasupat, Chen, Chaganty, Fan, Zhao, Lao, Lee, Juan, and Guu}]{gao2023rarr}
Luyu Gao, Zhuyun Dai, Panupong Pasupat, Anthony Chen, Arun~Tejasvi Chaganty, Yicheng Fan, Vincent Zhao, Ni~Lao, Hongrae Lee, Da-Cheng Juan, and Kelvin Guu. 2023.
\newblock \href {https://doi.org/10.18653/v1/2023.acl-long.910} {{RARR}: Researching and revising what language models say, using language models}.
\newblock In \emph{Proceedings of the 61st Annual Meeting of the Association for Computational Linguistics (Volume 1: Long Papers)}, pages 16477--16508, Toronto, Canada. Association for Computational Linguistics.

\bibitem[{Grattafiori et~al.(2024)Grattafiori, Dubey, Jauhri, Pandey, Kadian, Al-Dahle, Letman, Mathur, Schelten, Vaughan et~al.}]{grattafiori2024llama}
Aaron Grattafiori, Abhimanyu Dubey, Abhinav Jauhri, Abhinav Pandey, Abhishek Kadian, Ahmad Al-Dahle, Aiesha Letman, Akhil Mathur, Alan Schelten, Alex Vaughan, and 1 others. 2024.
\newblock The llama 3 herd of models.
\newblock \emph{arXiv preprint arXiv:2407.21783}.

\bibitem[{Gunjal and Durrett(2024)}]{gunjal2024molecular}
Anisha Gunjal and Greg Durrett. 2024.
\newblock \href {https://doi.org/10.18653/v1/2024.findings-emnlp.215} {Molecular facts: Desiderata for decontextualization in {LLM} fact verification}.
\newblock In \emph{Findings of the Association for Computational Linguistics: EMNLP 2024}, pages 3751--3768, Miami, Florida, USA. Association for Computational Linguistics.

\bibitem[{Honovich et~al.(2022)Honovich, Aharoni, Herzig, Taitelbaum, Kukliansy, Cohen, Scialom, Szpektor, Hassidim, and Matias}]{honovich2022true}
Or~Honovich, Roee Aharoni, Jonathan Herzig, Hagai Taitelbaum, Doron Kukliansy, Vered Cohen, Thomas Scialom, Idan Szpektor, Avinatan Hassidim, and Yossi Matias. 2022.
\newblock \href {https://doi.org/10.18653/v1/2022.naacl-main.287} {{TRUE}: Re-evaluating factual consistency evaluation}.
\newblock In \emph{Proceedings of the 2022 Conference of the North American Chapter of the Association for Computational Linguistics: Human Language Technologies}, pages 3905--3920, Seattle, United States. Association for Computational Linguistics.

\bibitem[{Hu et~al.(2025)Hu, Long, and Wang}]{hu2025decomp}
Qisheng Hu, Quanyu Long, and Wenya Wang. 2025.
\newblock \href {https://doi.org/10.18653/v1/2025.naacl-long.320} {Decomposition dilemmas: Does claim decomposition boost or burden fact-checking performance?}
\newblock In \emph{Proceedings of the 2025 Conference of the Nations of the Americas Chapter of the Association for Computational Linguistics: Human Language Technologies (Volume 1: Long Papers)}, pages 6313--6336, Albuquerque, New Mexico. Association for Computational Linguistics.

\bibitem[{Kamoi et~al.(2023)Kamoi, Goyal, Diego~Rodriguez, and Durrett}]{kamoi2023wice}
Ryo Kamoi, Tanya Goyal, Juan Diego~Rodriguez, and Greg Durrett. 2023.
\newblock \href {https://doi.org/10.18653/v1/2023.emnlp-main.470} {{W}i{CE}: Real-world entailment for claims in {W}ikipedia}.
\newblock In \emph{Proceedings of the 2023 Conference on Empirical Methods in Natural Language Processing}, pages 7561--7583, Singapore. Association for Computational Linguistics.

\bibitem[{Laban et~al.(2022)Laban, Schnabel, Bennett, and Hearst}]{laban2022summac}
Philippe Laban, Tobias Schnabel, Paul~N. Bennett, and Marti~A. Hearst. 2022.
\newblock \href {https://doi.org/10.1162/tacl_a_00453} {{S}umma{C}: Re-visiting {NLI}-based models for inconsistency detection in summarization}.
\newblock \emph{Transactions of the Association for Computational Linguistics}, 10:163--177.

\bibitem[{Laurer et~al.(2023)Laurer, Van~Atteveldt, Casas, and Welbers}]{laurer_less_2023}
Moritz Laurer, Wouter Van~Atteveldt, Andreu Casas, and Kasper Welbers. 2023.
\newblock \href {https://doi.org/10.1017/pan.2023.20} {Less {Annotating}, {More} {Classifying}: {Addressing} the {Data} {Scarcity} {Issue} of {Supervised} {Machine} {Learning} with {Deep} {Transfer} {Learning} and {BERT}-{NLI}}.
\newblock \emph{Political Analysis}, pages 1--33.

\bibitem[{Lu et~al.(2025)Lu, Ziems, Dang, and Jiang}]{lu2025optimizing}
Yining Lu, Noah Ziems, Hy~Dang, and Meng Jiang. 2025.
\newblock \href {https://doi.org/10.18653/v1/2025.acl-long.254} {Optimizing decomposition for optimal claim verification}.
\newblock In \emph{Proceedings of the 63rd Annual Meeting of the Association for Computational Linguistics (Volume 1: Long Papers)}, pages 5095--5114, Vienna, Austria. Association for Computational Linguistics.

\bibitem[{McCoy et~al.(2019)McCoy, Pavlick, and Linzen}]{mccoy2019right}
R.~Thomas McCoy, Ellie Pavlick, and Tal Linzen. 2019.
\newblock \href {https://doi.org/10.18653/v1/P19-1334} {Right for the wrong reasons: Diagnosing syntactic heuristics in natural language inference}.
\newblock In \emph{Proceedings of the 57th Annual Meeting of the Association for Computational Linguistics}, pages 3428--3448, Florence, Italy. Association for Computational Linguistics.

\bibitem[{McNemar(1947)}]{mcnemar1947note}
Quinn McNemar. 1947.
\newblock Note on the sampling error of the difference between correlated proportions or percentages.
\newblock \emph{Psychometrika}, 12(2):153--157.

\bibitem[{Min et~al.(2023)Min, Krishna, Lyu, Lewis, Yih, Koh, Iyyer, Zettlemoyer, and Hajishirzi}]{min2023factscore}
Sewon Min, Kalpesh Krishna, Xinxi Lyu, Mike Lewis, Wen-tau Yih, Pang Koh, Mohit Iyyer, Luke Zettlemoyer, and Hannaneh Hajishirzi. 2023.
\newblock \href {https://doi.org/10.18653/v1/2023.emnlp-main.741} {{FA}ct{S}core: Fine-grained atomic evaluation of factual precision in long form text generation}.
\newblock In \emph{Proceedings of the 2023 Conference on Empirical Methods in Natural Language Processing}, pages 12076--12100, Singapore. Association for Computational Linguistics.

\bibitem[{Pearson(1900)}]{pearson1900correlation}
Karl Pearson. 1900.
\newblock I. mathematical contributions to the theory of evolution.—vii. on the correlation of characters not quantitatively measurable.
\newblock \emph{Philosophical Transactions of the Royal Society of London. Series A, Containing Papers of a Mathematical or Physical Character}, 195(262-273):1--47.

\bibitem[{Popovi{\v{c}} and F{\"a}rber(2025)}]{popovic2025jedi}
Nicholas Popovi{\v{c}} and Michael F{\"a}rber. 2025.
\newblock \href {https://doi.org/10.18653/v1/2025.emnlp-main.1615} {Extractive fact decomposition for interpretable natural language inference in one forward pass}.
\newblock In \emph{Proceedings of the 2025 Conference on Empirical Methods in Natural Language Processing}, pages 31692--31705, Suzhou, China. Association for Computational Linguistics.

\bibitem[{Qi et~al.(2024)Qi, Sarti, Fern{\'a}ndez, and Bisazza}]{qi2024mirage}
Jirui Qi, Gabriele Sarti, Raquel Fern{\'a}ndez, and Arianna Bisazza. 2024.
\newblock \href {https://doi.org/10.18653/v1/2024.emnlp-main.347} {Model internals-based answer attribution for trustworthy retrieval-augmented generation}.
\newblock In \emph{Proceedings of the 2024 Conference on Empirical Methods in Natural Language Processing}, pages 6037--6053, Miami, Florida, USA. Association for Computational Linguistics.

\bibitem[{Rashkin et~al.(2023)Rashkin, Nikolaev, Lamm, Aroyo, Collins, Das, Petrov, Tomar, Turc, and Reitter}]{rashkin2021ais}
Hannah Rashkin, Vitaly Nikolaev, Matthew Lamm, Lora Aroyo, Michael Collins, Dipanjan Das, Slav Petrov, Gaurav~Singh Tomar, Iulia Turc, and David Reitter. 2023.
\newblock \href {https://doi.org/10.1162/coli_a_00486} {Measuring attribution in natural language generation models}.
\newblock \emph{Computational Linguistics}, 49(4):777--840.

\bibitem[{Seo et~al.(2025)Seo, Han, Jung, Newman, Lim, Lee, Lu, Choi, and Yu}]{seo2025verifying}
Wooseok Seo, Seungju Han, Jaehun Jung, Benjamin Newman, Seungwon Lim, Seungbeen Lee, Ximing Lu, Yejin Choi, and Youngjae Yu. 2025.
\newblock Verifying the verifiers: Unveiling pitfalls and potentials in fact verifiers.

\bibitem[{Sharma et~al.(2024)Sharma, Tong, Korbak, Duvenaud, Askell, Bowman, DURMUS, Hatfield-Dodds, Johnston, Kravec, Maxwell, McCandlish, Ndousse, Rausch, Schiefer, Yan, Zhang, and Perez}]{sharma2024sycophancy}
Mrinank Sharma, Meg Tong, Tomek Korbak, David Duvenaud, Amanda Askell, Sam Bowman, Esin DURMUS, Zac Hatfield-Dodds, Scott Johnston, Shauna Kravec, Timothy Maxwell, Sam McCandlish, Kamal Ndousse, Oliver Rausch, Nicholas Schiefer, Da~Yan, Miranda Zhang, and Ethan Perez. 2024.
\newblock \href {https://proceedings.iclr.cc/paper_files/paper/2024/file/0105f7972202c1d4fb817da9f21a9663-Paper-Conference.pdf} {Towards understanding sycophancy in language models}.
\newblock In \emph{International Conference on Learning Representations}, volume 2024, pages 110--144.

\bibitem[{Spearman(1904)}]{spearman1904proof}
C.~Spearman. 1904.
\newblock \href {http://www.jstor.org/stable/1412159} {The proof and measurement of association between two things}.
\newblock \emph{The American Journal of Psychology}, 15(1):72--101.

\bibitem[{Tang et~al.(2024)Tang, Laban, and Durrett}]{tang2024minicheck}
Liyan Tang, Philippe Laban, and Greg Durrett. 2024.
\newblock \href {https://doi.org/10.18653/v1/2024.emnlp-main.499} {{M}ini{C}heck: Efficient fact-checking of {LLM}s on grounding documents}.
\newblock In \emph{Proceedings of the 2024 Conference on Empirical Methods in Natural Language Processing}, pages 8818--8847, Miami, Florida, USA. Association for Computational Linguistics.

\bibitem[{Tang et~al.(2025)Tang, Wang, and Tung}]{tang2025tracer}
Yixuan Tang, Jincheng Wang, and Anthony Kum~Hoe Tung. 2025.
\newblock \href {https://doi.org/10.18653/v1/2025.emnlp-main.1724} {The missing parts: Augmenting fact verification with half truth detection}.
\newblock In \emph{Proceedings of the 2025 Conference on Empirical Methods in Natural Language Processing}, pages 33979--33996, Suzhou, China. Association for Computational Linguistics.

\bibitem[{Tang et~al.(2026)Tang, Zhang, Feng, and Tung}]{debating2025radar}
Yixuan Tang, Yirui Zhang, Hang Feng, and Anthony K.~H. Tung. 2026.
\newblock \href {https://arxiv.org/abs/2604.19005} {Debating the unspoken: Role-anchored multi-agent reasoning for half-truth detection}.

\bibitem[{Team et~al.(2024)Team, Riviere, Pathak, Sessa, Hardin, Bhupatiraju, Hussenot, Mesnard, Shahriari, Ram{\'e} et~al.}]{team2024gemma}
Gemma Team, Morgane Riviere, Shreya Pathak, Pier~Giuseppe Sessa, Cassidy Hardin, Surya Bhupatiraju, L{\'e}onard Hussenot, Thomas Mesnard, Bobak Shahriari, Alexandre Ram{\'e}, and 1 others. 2024.
\newblock Gemma 2: Improving open language models at a practical size.
\newblock \emph{arXiv preprint arXiv:2408.00118}.

\bibitem[{Thorne et~al.(2018)Thorne, Vlachos, Christodoulopoulos, and Mittal}]{thorne2018fever}
James Thorne, Andreas Vlachos, Christos Christodoulopoulos, and Arpit Mittal. 2018.
\newblock \href {https://doi.org/10.18653/v1/N18-1074} {{FEVER}: a large-scale dataset for fact extraction and {VER}ification}.
\newblock In \emph{Proceedings of the 2018 Conference of the North {A}merican Chapter of the Association for Computational Linguistics: Human Language Technologies, Volume 1 (Long Papers)}, pages 809--819, New Orleans, Louisiana. Association for Computational Linguistics.

\bibitem[{Tonglet et~al.(2024)Tonglet, Moens, and Gurevych}]{tonglet2024image}
Jonathan Tonglet, Marie-Francine Moens, and Iryna Gurevych. 2024.
\newblock \href {https://doi.org/10.18653/v1/2024.emnlp-main.448} {``image, tell me your story!'' predicting the original meta-context of visual misinformation}.
\newblock pages 7845--7864.

\bibitem[{Tonglet et~al.(2025)Tonglet, Thiem, and Gurevych}]{tonglet2025cove}
Jonathan Tonglet, Gabriel Thiem, and Iryna Gurevych. 2025.
\newblock \href {https://doi.org/10.18653/v1/2025.naacl-long.102} {{COVE}: {CO}ntext and {VE}racity prediction for out-of-context images}.
\newblock In \emph{Proceedings of the 2025 Conference of the Nations of the Americas Chapter of the Association for Computational Linguistics: Human Language Technologies (Volume 1: Long Papers)}, pages 2029--2049, Albuquerque, New Mexico. Association for Computational Linguistics.

\bibitem[{Toulmin(2003)}]{toulmin1958uses}
Stephen~E Toulmin. 2003.
\newblock \emph{The uses of argument}.
\newblock Cambridge university press.

\bibitem[{Urbani(2020)}]{urbani2020verifying}
S.~Urbani. 2020.
\newblock \href {https://firstdraftnews.org/articles/verifying-online-information-the-absolute-essentials/} {Verifying online information}.
\newblock \emph{Essential Guides}.
\newblock Published in 2022.

\bibitem[{Wadden et~al.(2020)Wadden, Lin, Lo, Wang, van Zuylen, Cohan, and Hajishirzi}]{wadden2020scifact}
David Wadden, Shanchuan Lin, Kyle Lo, Lucy~Lu Wang, Madeleine van Zuylen, Arman Cohan, and Hannaneh Hajishirzi. 2020.
\newblock \href {https://doi.org/10.18653/v1/2020.emnlp-main.609} {Fact or fiction: Verifying scientific claims}.
\newblock In \emph{Proceedings of the 2020 Conference on Empirical Methods in Natural Language Processing (EMNLP)}, pages 7534--7550, Online. Association for Computational Linguistics.

\bibitem[{Wanner et~al.(2025)Wanner, Van~Durme, and Dredze}]{wanner2025dnd}
Miriam Wanner, Benjamin Van~Durme, and Mark Dredze. 2025.
\newblock \href {https://doi.org/10.18653/v1/2025.emnlp-main.1205} {{D}n{DS}core: Decontextualization and decomposition for factuality verification in long-form text generation}.
\newblock In \emph{Proceedings of the 2025 Conference on Empirical Methods in Natural Language Processing}, pages 23609--23626, Suzhou, China. Association for Computational Linguistics.

\bibitem[{Wei et~al.(2024)Wei, Yang, Song, Lu, Hu, Huang, Tran, Peng, Liu, Huang, Du, and Le}]{wei2024safe}
Jerry Wei, Chengrun Yang, Xinying Song, Yifeng Lu, Nathan Hu, Jie Huang, Dustin Tran, Daiyi Peng, Ruibo Liu, Da~Huang, Cosmo Du, and Quoc~V. Le. 2024.
\newblock Long-form factuality in large language models.
\newblock In \emph{Proceedings of the 38th International Conference on Neural Information Processing Systems}, NIPS '24, Red Hook, NY, USA. Curran Associates Inc.

\bibitem[{Yang et~al.(2025)Yang, Li, Yang, Zhang, Hui, Zheng, Yu, Gao, Huang, Lv et~al.}]{qwen3technicalreport}
An~Yang, Anfeng Li, Baosong Yang, Beichen Zhang, Binyuan Hui, Bo~Zheng, Bowen Yu, Chang Gao, Chengen Huang, Chenxu Lv, and 1 others. 2025.
\newblock Qwen3 technical report.

\end{thebibliography}

\appendix

\setcounter{section}{0}
\renewcommand{\thesection}{\Alph{section}}
\setcounter{table}{0}
\renewcommand{\thetable}{S\arabic{table}}
\setcounter{figure}{0}
\renewcommand{\thefigure}{S\arabic{figure}}

\section{SIFT Algorithm}
\label{app:sift}
\begin{algorithm}[h]
\scriptsize
\caption{SIFT with claim-conditioned re-scoring}
\label{alg:sift}
\begin{algorithmic}[1]
\Require claim $c$, evidence pool $\mathcal{E}$, backbone $f$
\State $\mathcal{U} \gets \textsc{ActiveUnits}(f,c)$
\For{$u \in \mathcal{U}$, $e \in \mathcal{E}$}
  \State $(S[e,u],K[e,u],\sigma[e,u]) \gets \textsc{Extract}(f,c,u,e)$
\EndFor
\State $(\hat{y},\hat{w}) \gets \textsc{Aggregate}(S,K,\sigma)$ \Comment{base 5W1H verdict}
\State $(a,\sigma^{\star}) \gets \textsc{ReScore}(f,c,\mathcal{E})$ \Comment{$a\in\{\textsc{Sup},\textsc{Con},\textsc{None}\}$}
\If{$a=\textsc{Sup}$ \textbf{and} $\textsc{Grounded}(\sigma^{\star})$ \textbf{and} $\hat{y}\neq\textsc{Supports}$}
  \State $(\hat{y},\hat{w}) \gets (\textsc{Supports},\,\sigma^{\star})$ \Comment{promote}
\ElsIf{$a=\textsc{Con}$ \textbf{and} $\textsc{Grounded}(\sigma^{\star})$ \textbf{and} $\hat{y}=\textsc{Supports}$}
  \State $\hat{y} \gets \textsc{Refutes}$ \Comment{demote}
\EndIf
\State \Return $(\hat{y},\hat{w})$
\end{algorithmic}
\end{algorithm}

Lines~1--4 are the 5W1H decomposition stage: the backbone selects the active claim units and extracts verbatim support and contradiction spans, forming the coverage matrix. Line~5 aggregates that matrix into a base verdict. Line~6 is claim-conditioned re-scoring: conditioned on the \emph{full} claim $c$, the backbone re-checks whether the evidence contains a grounded span that explicitly licenses or contradicts $c$ as a whole. Lines~7--11 apply that check, and it is bidirectional---a base non-\textsc{Supports} verdict is promoted when a grounded span explicitly licenses $c$, and a base \textsc{Supports} verdict is demoted when one explicitly contradicts it; otherwise the base verdict stands. Naive 5W1H is the same procedure without the re-scoring step (lines~6--11); it is the failure condition, not a separate system. \textsc{Extract} returns verbatim substrings, and \textsc{Grounded} requires the cited span $\sigma^{\star}$ to occur in its cited evidence. The released implementation runs \textsc{ReScore} as two passes---a certificate-style grounding pass and a claim-conditioned verdict pass---which Algorithm~\ref{alg:sift} summarises as a single step.

\section{Late-Stage Verifier}
\label{app:thresholds}

\begin{algorithm}[h]
\scriptsize
\caption{Design~B deterministic rule}
\label{alg:designb}
\begin{algorithmic}[1]
\Require SIFT verdict $v_A$, SIFT sublabel $r_A$, sweep entailment $s_E^{\mathrm{sweep}}$, sweep contradiction $s_C^{\mathrm{sweep}}$, SIFT warrant entailment $s_E^{\mathrm{SIFT}}$, SIFT warrant contradiction $s_C^{\mathrm{SIFT}}$, thresholds $\tau$
\State $s_E \gets \max(s_E^{\mathrm{sweep}},s_E^{\mathrm{SIFT}})$
\State $s_C \gets \max(s_C^{\mathrm{sweep}},s_C^{\mathrm{SIFT}})$
\If{$v_A = \mathrm{SUPPORTS}$}
  \State $f \gets r_A$ is a fragile support sublabel
  \If{$\tau_{\mathrm{fragileOnly}}$ and not $f$}
    \State \Return \textsc{Keep}
  \EndIf
  \If{$(s_E^{\mathrm{sweep}} < \tau_{\mathrm{vetoE}}$ or $s_C > \tau_{\mathrm{vetoC}})$ and $s_E < \tau_{\mathrm{protect}}$}
    \State \Return \textsc{Veto} \Comment{$\rightarrow \mathrm{NEI}$}
  \EndIf
  \State \Return \textsc{Keep}
\Else
  \State $m \gets r_A$ is a pre-declared rescue-eligible sublabel
  \If{$m$ and $s_E > \tau_{\mathrm{rescueE}}$ and $s_C < \tau_{\mathrm{rescueC}}$}
    \State \Return \textsc{Rescue} \Comment{$\rightarrow \mathrm{SUPPORTS}$}
  \EndIf
  \State \Return \textsc{Keep}
\EndIf
\end{algorithmic}
\end{algorithm}

\paragraph{Threshold selection.} Design~B uses a finite validation grid over $\tau_{\mathrm{vetoE}}\in\{0.00,0.20,0.40\}$, $\tau_{\mathrm{vetoC}}\in\{0.85,0.90,0.95,0.98\}$, $\tau_{\mathrm{protect}}\in\{0.80,0.90,0.95,0.98\}$, $\tau_{\mathrm{rescueE}}\in\{0.85,0.90,0.95,0.98\}$, $\tau_{\mathrm{rescueC}}\in\{0.50,0.80,0.95\}$, and $\tau_{\mathrm{fragileOnly}}\in\{\mathrm{true},\mathrm{false}\}$. The tuple is selected on the 30\% validation split, pooled by dataset across the primary backbones (Qwen3-4B, Llama3.1-8B, Gemma2-9B), then frozen for held-out evaluation. Per-cell selected tuples and action traces are released with the code and data.

\paragraph{Global-tuple sensitivity.} A single global tuple selected across datasets matches the per-dataset top-1 operating point on FEVER and SciFact, improves DP by 0.2pp, and costs 2.0pp on 5PILS while matching or improving WSP everywhere. This is consistent with treating Design~B as a stability layer that benefits from calibration data rather than as a universal rule.

\section{Design~A Prompts}
\label{app:design-a-prompts}

Design~A is implemented as two prompts. The \emph{rescue-only} variant calls only the rescue prompt on non-\textsc{Supports} verdicts; the \emph{two-sided} variant additionally calls the veto prompt on fragile \textsc{Supports} verdicts. Both prompts are wrapped by a strict format instruction and parsed by a JSON-first decoder with a regex fallback that picks the first occurrence of an allowed decision token if JSON parsing fails. On the repaired Llama3.1-8B run, the resulting invalid-decision rate is $0\%$.

\paragraph{Format wrapper (both variants).}
\begin{quote}\ttfamily\scriptsize
Return ONLY one minified JSON object. No markdown. No explanation before or after JSON. The first character must be \{ and the last character must be \}. Use exactly one allowed decision token from the output\_schema.
\end{quote}

\paragraph{Veto prompt (two-sided only).} Issued for SIFT \textsc{Supports} verdicts carrying a fragility sub-label. The JSON payload contains the claim, the SIFT verdict object, its connected-evidence text, and the NLI sweep diagnostics. The decision token is \texttt{KEEP\_SUPPORTS} or \texttt{DOWNGRADE\_NEI}, governed by these rules (verbatim from the implementation):
\begin{quote}\ttfamily\scriptsize
Return JSON only.\\
First write a 1--2 sentence rationale grounded in specific document alignment or conflict.\\
Do not downgrade merely because support comes from multiple documents; multi-document support is valid when entity, event, place, and time align.\\
Choose KEEP\_SUPPORTS only if the evidence supports the full claim as one coherent event.\\
Choose DOWNGRADE\_NEI if the support is patchwork, cross-event, cross-date, or only topically related.\\
Do not use outside knowledge.\\
Be conservative only when there is a concrete mismatch; otherwise keep the ADMIT SUPPORTS verdict.
\end{quote}

\paragraph{Rescue prompt (both variants).} Issued for non-\textsc{Supports} verdicts carrying a rescue sub-label. The payload contains the claim, the SIFT verdict, the top-entailment and top-contradiction NLI windows, and nearby evidence text. The decision token is \texttt{PROMOTE\_SUPPORTS} or \texttt{KEEP\_NON\_SUPPORTS}, governed by:
\begin{quote}\ttfamily\scriptsize
Return JSON only.\\
First write a rescue\_rationale explaining the specific entity/action/location/time fit.\\
Choose PROMOTE\_SUPPORTS only if the candidate evidence explicitly supports the full claim.\\
Choose KEEP\_NON\_SUPPORTS if any important entity, action, location, time, or image-use detail is missing.\\
Do not promote when evidence is merely related.\\
Do not use outside knowledge.
\end{quote}

The full prompt builders, including the JSON payload schemas and the regex-fallback decoder, are released with the code.

\vspace{-10pt}
\section{Design~B Action Counts and No-op Cells}
\label{app:designb-actions}

The Design~B rule (Appendix~\ref{app:thresholds}) is applied with one global threshold tuple to every held-out cell. Whether the rule acts on a given cell depends on two structural factors that are checked \emph{before} any threshold: a \emph{dataset factor}---SIFT's sub-label distribution, which determines how many items carry a fragility or rescue marker---and a \emph{model$\times$dataset factor}---the NLI score distribution on those marked items, which determines whether the threshold conditions are met. Tables~\ref{tab:designb-veto-actions} and~\ref{tab:designb-rescue-actions} decompose action counts along both axes for every primary cell on the same 70\% held-out test split as Table~\ref{tab:verifier}. Fragility markers are \textsc{Patchwork}, \textsc{Direct\_NonWho}, \textsc{Action\_Anchor\_Rescue}; rescue markers are \textsc{Missing\_Evidence}, \textsc{Empty\_Matrix}, \textsc{No\_Connected\_Evidence}, \textsc{False\_\{Location,Time,Action\}}, \textsc{Disputed\_\{Action,Where,When\}}, \textsc{Out\_of\_Context\_Semantic}.

\begin{table}[t]
\scriptsize
\setlength{\tabcolsep}{4pt}
\begin{tabular}{@{} l l r r r r r @{}}
\toprule
\textbf{Model} & \textbf{Dataset} & \textbf{SIFT-S} & \textbf{Frag.} & \textbf{Cond.} & \textbf{Prot.} & \textbf{VETO} \\
\midrule
Qwen3-4B    & 5PILS   & 337 & 295 & 206 & 177 & 115 \\
Qwen3-4B    & DP    & 150 &  73 &  54 &  27 &  44 \\
Qwen3-4B    & FEVER   & 339 &   0 &   0 &   0 &   0 \\
Qwen3-4B    & SciFact &  85 &   0 &   0 &   0 &   0 \\
\addlinespace
Llama3.1-8B & 5PILS   & 331 & 315 & 245 & 216 &  97 \\
Llama3.1-8B & DP    & 142 & 119 &  79 &  82 &  34 \\
Llama3.1-8B & FEVER   & 368 &   0 &   0 &   0 &   0 \\
Llama3.1-8B & SciFact &  50 &   0 &   0 &   0 &   0 \\
\addlinespace
Gemma2-9B   & 5PILS   & 462 & 402 & 320 & 278 & 120 \\
Gemma2-9B   & DP    & 365 & 313 & 236 & 166 & 138 \\
Gemma2-9B   & FEVER   & 485 &   0 &   0 &   0 &   0 \\
Gemma2-9B   & SciFact & 102 &   0 &   0 &   0 &   0 \\
\bottomrule
\end{tabular}
\caption{\textbf{Veto path action counts.} \textbf{SIFT-S}: predicted-\textsc{Supports} on held-out test. \textbf{Frag.}: items carrying a fragility sub-label marker. \textbf{Cond.}: fragile items satisfying veto condition ($s_{\text{sweep}}<0.4$ or $k_{\max}>0.85$). \textbf{Prot.}: fragile items with $s_{\max}\ge 0.9$ (protected). \textbf{VETO}: vetoes that fire (Cond.\ and not Prot.). Threshold values ($\tau_{\mathrm{vetoE}}{=}0.4$, $\tau_{\mathrm{vetoC}}{=}0.85$, $\tau_{\mathrm{protect}}{=}0.9$) are the selected global tuple from Appendix~\ref{app:thresholds}. FEVER and SciFact produce zero fragile SIFT-\textsc{Supports} for every model, so the veto path is structurally inactive on the gold-evidence datasets.}
\label{tab:designb-veto-actions}
\end{table}

\begin{table}[t]
\centering
\scriptsize
\setlength{\tabcolsep}{4pt}
\begin{tabular*}{\columnwidth}{@{} l l r r r r @{}}
\toprule
\textbf{Model} & \textbf{Dataset} & \textbf{non-S} & \textbf{Resc.\ elig.} & $\max(s_{\max})$ & \textbf{RESCUE} \\
\midrule
Qwen3-4B    & 5PILS   & 379 & 361 & 1.00 & 27 \\
Qwen3-4B    & DP    & 266 & 251 & 1.00 & 51 \\
Qwen3-4B    & FEVER   & 367 & 245 & 1.00 & 27 \\
Qwen3-4B    & SciFact &  46 &  37 & 1.00 &  6 \\
\addlinespace
Llama3.1-8B & 5PILS   & 385 & 375 & 1.00 & 28 \\
Llama3.1-8B & DP    & 274 & 260 & 1.00 & 38 \\
Llama3.1-8B & FEVER   & 338 & 301 & 1.00 & 59 \\
Llama3.1-8B & SciFact &  81 &  81 & 1.00 & 15 \\
\addlinespace
Gemma2-9B   & 5PILS   & 254 & 241 & 1.00 &  7 \\
Gemma2-9B   & DP    &  51 &  48 & 1.00 &  4 \\
\rowcolor{crowhi}
Gemma2-9B   & FEVER   & 221 & 122 & \textbf{0.71} & \textbf{0} \\
Gemma2-9B   & SciFact &  29 &  22 & 0.95 &  1 \\
\bottomrule
\end{tabular*}
\caption{\textbf{Rescue path action counts.} \textbf{non-S}: non-\textsc{Supports} verdicts on held-out test. \textbf{Resc.\ elig.}: items carrying a rescue marker. $\max(s_{\max})$: maximum entailment over the warrant premise and the NLI sweep windows, taken over rescue-eligible items. \textbf{RESCUE}: rescues that fire ($s_{\max}{>}\tau_{\mathrm{rescueE}}{=}0.85$ and $k_{\max}{<}\tau_{\mathrm{rescueC}}{=}0.8$);  thresholds are the selected global tuple, Appendix~\ref{app:thresholds}. The shaded row is the Design~B = SIFT cell: Gemma's rescue-eligible non-\textsc{Supports} verdicts on FEVER are all NLI-confirmed (no item clears $0.85$), so no candidate exists for the rule to rescue.}
\label{tab:designb-rescue-actions}
\end{table}

\paragraph{Veto silence on FEVER and SciFact is structural.} On both gold-evidence datasets, every model produces zero fragile SIFT-\textsc{Supports}: SIFT cannot stitch a patchwork warrant when the evidence pool is a single human-curated gold sentence. The veto path is therefore inactive on all six (model, dataset) pairs in $\{$Qwen3-4B, Llama3.1-8B, Gemma2-9B$\}\times\{$FEVER, SciFact$\}$, and the global tuple has nothing to demote there.

\paragraph{Rescue silence on Gemma2-9B/FEVER is model$\times$dataset specific.} Among Gemma's 122 rescue-eligible non-\textsc{Supports} items on FEVER, the maximum NLI entailment over both the warrant premise and the three-sentence window sweep is $0.71$, below the $0.85$ rescue threshold. On the same dataset, Qwen3-4B reaches $1.00$ (27 rescues fire), and Llama3.1-8B reaches $1.00$ (59 rescues). Gemma's SIFT non-\textsc{Supports} verdicts on FEVER are corroborated by the independent NLI sweep, which contains no high-entailment evidence for the rescue path to act on.

\paragraph{Cross-dataset threshold transfer.} The global tuple is a stabiliser: it acts on disagreement between the SIFT verdict and the NLI sweep, and stands down on agreement. The cells where it intervenes most heavily---5PILS and DP on every backbone---are the regimes where rigid 5W1H actually produces patchwork warrants (fragile counts 73--402). The cells where it stands down---FEVER and SciFact for veto on every backbone, Gemma2-9B/FEVER for both paths---are those where SIFT already converges with the independent NLI sweep. The cost of using a single tuple everywhere is that no cell receives per-cell optimisation; the benefit, as shown in Table~\ref{tab:verifier}, is that no cell is degraded.

\section{External Audits}
\label{app:audits}

\paragraph{Gemini LLM-PPI++.} The blind audit uses only the claim and exact NLI premise, hiding system identity and NLI scores. We audit 250 items in two distinct strata: \textbf{(i)}~100 stratified-random Qwen3-4B predicted-\textsc{Supports} items pooled across Direct, SIFT, and Design~B, used for the PPI++ population estimate over the full $n_{\text{pop}}{=}3620$ predicted-\textsc{Supports} pool; and \textbf{(ii)}~150 targeted delta-cell items where Direct, SIFT, and Design~B disagree, reported separately as a stress test (not pooled into the population estimate). PPI++ estimates \citep{angelopoulos2024ppipp}:
\begin{equation}
  \hat{\theta}^{\mathrm{PPI++}} = \frac{1}{n}\sum_{i=1}^{n}(Y_i-\lambda\hat{Y}_i) + \frac{\lambda}{N}\sum_{j=1}^{N}\hat{Y}_j,
\end{equation}
with $\lambda$ selected by \texttt{ppi-py} to minimize CI width. On the 100-item random stratum: pooled admissibility $58.2$ [51.3, 65.1]; $\phi=0.720$; inter-run binary $\alpha=0.912$; six-label diagnostic $\alpha=0.876$.

\paragraph{Per-dataset estimates (random subsets).} 5PILS: $48.6$ [36.9, 60.2]. DP: $45.0$ [29.3, 60.7]. FEVER: $77.2$ [64.5, 89.9]. SciFact: $48.3$ [34.4, 62.2]. The FEVER level is closest to the NLI training distribution; multi-evidence and scientific claims sit lower. Per-dataset subsets are small, so agreement with WSP is reported only at the pooled level.

\section{Base Rates for WSP-Gold Calibration}
  \label{app:base-rates}
  The precision column in Table~\ref{tab:wsp-gold} is read against a base rate: the prior probability that a
  predicted-\textsc{Supports} verdict is gold-admissible before any WSP filtering. Concretely,
  \[
  \mathrm{base\_rate} = \frac{\bigl|\{i \in S : y^{\mathrm{gold}}_i = \textsc{Supports}\}\bigr|}{|S|},
  \]
  where $S$ is the pool of predicted-\textsc{Supports} verdicts on FEVER and SciFact across the three primary backbones (Qwen3-4B, Llama3.1-8B, Gemma2-9B), and $y^{\mathrm{gold}}_i$ is the gold verdict from the source dataset. At SIFT stage, $1550/2036$ predicted-\textsc{Supports} are gold-admissible, giving base rate $0.76$. At Direct stage, the predicted-\textsc{Supports} pool is smaller and more selective: Direct emits SUPPORTS less often than SIFT, but each call is more often correct, yielding base rate $0.88$ on $1667$ items. WSP's precision ($0.98$ at SIFT, $0.98$ at Direct; Table~\ref{tab:wsp-gold}) sits well above both  base rates, so the filtering gain holds regardless of upstream stage.

\section{Paired McNemar Significance}
\label{app:mcnemar}

\begin{table}[h]
\centering
\scriptsize
\setlength{\tabcolsep}{3pt}
\begin{tabular}{@{}llrrrr@{}}
\toprule
& & \multicolumn{4}{c}{\textbf{Design B vs SIFT}} \\
\cmidrule(lr){3-6}
\textbf{Model} & \textbf{Dataset} & $\Delta$pp & $b$ & $c$ & $p_{\mathrm{BH}}$ \\
\midrule
Qwen3-4B    & 5PILS   & $+0.0$  & 71 & 71 & 1.000 \\
Qwen3-4B    & DP    & $+3.6$  & 55 & 40 & 0.249 \\
Qwen3-4B    & FEVER   & $+3.5$  & 26 &  1 & \textbf{$<\!10^{-4}$} \\
Qwen3-4B    & SciFact & $+4.6$  &  6 &  0 & 0.100 \\
Llama3.1-8B & 5PILS   & $+3.2$  & 74 & 51 & 0.130 \\
Llama3.1-8B & DP    & $+4.8$  & 46 & 26 & 0.098 \\
Llama3.1-8B & FEVER   & $+8.1$  & 58 &  1 & \textbf{$<\!10^{-4}$} \\
Llama3.1-8B & SciFact & $+11.4$ & 15 &  0 & \textbf{0.0003} \\
Gemma2-9B   & 5PILS   & $+2.4$  & 72 & 55 & 0.249 \\
Gemma2-9B   & DP    & $-3.8$  & 63 & 79 & 0.277 \\
Gemma2-9B   & FEVER   & $+0.0$  &  0 &  0 & 1.000 \\
Gemma2-9B   & SciFact & $+0.8$  &  1 &  0 & 1.000 \\
Qwen3-14B   & 5PILS   & $-1.7$  & 18 & 30 & 0.249 \\
Qwen3-14B   & DP    & $-0.7$  & 15 & 18 & 0.896 \\
Qwen3-14B   & FEVER   & $+0.7$  &  7 &  2 & 0.261 \\
Qwen3-14B   & SciFact & $+3.1$  &  4 &  0 & 0.249 \\
\bottomrule
\end{tabular}
\caption{Two-sided exact McNemar tests with BH adjustment over the 16 cells. Bold values are significant at $\alpha=0.05$. Three BH-significant gains; no significant losses.}
\label{tab:mcnemar-app}
\end{table}

\section{Cross-Backbone WSP under MiniCheck}
\label{app:minicheck-app}

\begin{table}[h]
\centering
\resizebox{\columnwidth}{!}{%
\begin{tabular}{@{}llrrrrrr@{}}
\toprule
& & \multicolumn{3}{c}{\textbf{SIFT}} & \multicolumn{3}{c}{\textbf{Design B}} \\
\cmidrule(lr){3-5}\cmidrule(lr){6-8}
\textbf{Model} & \textbf{Dataset} & $n_{\text{S}}$ & DeB-WSP & MC-WSP & $n_{\text{S}}$ & DeB-WSP & MC-WSP \\
\midrule
Qwen3-4B    & 5PILS   & 337 & 55.5 & 46.0 & 249 & 77.5 & 61.5 \\
Qwen3-4B    & DP    & 150 & 42.7 & 38.0 & 157 & 63.7 & 52.2 \\
Qwen3-4B    & FEVER   & 339 & 75.5 & 62.5 & 366 & 75.1 & 61.5 \\
Qwen3-4B    & SciFact &  85 & 48.2 & 47.1 &  91 & 50.5 & 48.4 \\
Llama3.1-8B & 5PILS   & 331 & 65.9 & 57.4 & 262 & 86.6 & 70.6 \\
Llama3.1-8B & DP    & 142 & 67.6 & 59.9 & 146 & 81.5 & 71.2 \\
Llama3.1-8B & FEVER   & 368 & 64.4 & 53.0 & 427 & 67.0 & 54.6 \\
Llama3.1-8B & SciFact &  50 & 58.0 & 58.0 &  65 & 63.1 & 56.9 \\
Gemma2-9B   & 5PILS   & 462 & 55.6 & 50.9 & 349 & 71.4 & 60.7 \\
Gemma2-9B   & DP    & 365 & 48.8 & 47.7 & 231 & 68.0 & 64.5 \\
Gemma2-9B   & FEVER   & 485 & 60.2 & 48.3 & 485 & 60.2 & 48.3 \\
Gemma2-9B   & SciFact & 102 & 46.1 & 47.1 & 103 & 46.6 & 47.6 \\
Qwen3-14B   & 5PILS   & 315 & 63.8 & 56.8 & 299 & 69.6 & 58.5 \\
Qwen3-14B   & DP    & 229 & 60.3 & 56.8 & 232 & 62.1 & 57.3 \\
Qwen3-14B   & FEVER   & 354 & 79.1 & 63.0 & 363 & 79.6 & 63.4 \\
Qwen3-14B   & SciFact &  65 & 64.6 & 63.1 &  69 & 66.7 & 63.8 \\
\bottomrule
\end{tabular}
}
\caption{Cross-backbone WSP under MiniCheck. $n_{\text{S}}$ is the predicted-\textsc{Supports} count on held-out test. Under MiniCheck, Design~B exceeds SIFT in 13 of 16 cells, ties in 1, and has 2 small reversals.}
\label{tab:minicheck-app}
\end{table}

Per-dataset rank correlations between DeB-WSP and MC-WSP: $\rho=1.00$ on 5PILS, $\rho=0.93$ on DP, $\rho=1.00$ on FEVER, $\rho=0.93$ on SciFact. Global $\rho=0.89$ over 32 cells.

\section{Atomic-Decomposition Robustness Check}
\label{app:atomic}

To test whether the decomposition-fragmentation dilemma is specific to the 5W1H schema, we re-ran the pipeline with atomic-claim decomposition (FActScore/WiCE family) as the decomposition stage: a claim is split into a variable-length set of atomic facts, each fact is verified against the evidence pool, and the verdict is a conjunctive aggregate. \emph{atomic + re-score} additionally re-verifies each atomic fact with the full claim as context---an analogue of SIFT's claim-conditioned re-scoring---before the conjunctive aggregate. All cells use the primary backbones and the held-out protocol of the main experiments; \emph{Direct} and \emph{naive 5W1H} are repeated from Table~\ref{tab:decomp-repair} for reference.

\paragraph{Implementation.} The atomic baseline is an in-house reimplementation of the FActScore/WiCE prompting paradigm (decompose-then-verify), not a call to the released FActScore or WiCE codebases. Decomposition and verification both run through the \emph{same} LLM backbone and decoding settings as SIFT, with three verification variants (naive per-fact, claim-conditioned, sibling-aware control) that mirror the corresponding SIFT stages. Holding the inference stack constant ensures any accuracy or WSP gap between atomic and 5W1H is attributable to the decomposition scheme rather than to differences in model, sampling, or output parsing. The decomposition prompt requires every name, number, date, place, qualifier, and modifier in the claim to appear in some atomic fact, with no information added or dropped; parse failures are treated as a single-fact claim. The complete prompts will be released with the code upon publication.

Two findings stand out (Table~\ref{tab:atomic-app}). First, atomic decomposition does not reproduce the largest naive-5W1H accuracy losses: its worst regression against direct prompting is $-7.4$ points, against $-27.6$ for naive 5W1H. Because atomic decomposition is granularity-adaptive---it splits a claim into as many facts as the claim contains rather than forcing a fixed six-slot frame---it does not manufacture the underspecified units that fragment short claims. This locates the dilemma in rigid, forced slotting, not in decomposition as such. Second, claim-conditioned re-scoring transfers to this scheme, but its behaviour differs: on an already-adaptive scheme, it raises predicted-\textsc{Supports} volume, so accuracy edges up in most cells while WSP declines in 11 of 12. This is an operating-point shift, not the two-axis repair that re-scoring produces, where 5W1H genuinely fragmented the claim. This is consistent with the \emph{molecular facts} view \citep{gunjal2024molecular}: fully atomic facts trade decontextuality against minimality, so neither extreme of the granularity spectrum is ideal. We retain 5W1H as the decomposition scheme because its typed, fixed-arity coverage matrix is required by the deterministic Design~B rule and the failure taxonomy, which a variable-length atomic fact list does not support; we do not claim it is the most accurate decomposition.

\begin{table}[t]
\centering
\resizebox{\columnwidth}{!}{%
\begin{tabular}{@{}llcccccccc@{}}
\toprule
 & & \multicolumn{2}{c}{\textbf{5PILS}} & \multicolumn{2}{c}{\textbf{DP}} & \multicolumn{2}{c}{\textbf{FEVER}} & \multicolumn{2}{c}{\textbf{SciFact}} \\
\cmidrule(lr){3-4}\cmidrule(lr){5-6}\cmidrule(lr){7-8}\cmidrule(lr){9-10}
\textbf{Model} & \textbf{Stage} & Acc & WSP & Acc & WSP & Acc & WSP & Acc & WSP \\
\midrule
\multirow{4}{*}{Qwen3-4B}
 & Direct               & 71.8 & 50.2 & 57.9 & 45.9 & 88.2 & 68.7 & 69.7 & 48.1 \\
 & naive 5W1H           & 71.1 & 54.4 & 41.5 & 29.7 & 64.6 & 56.1 & 71.3 & 43.9 \\
 & atomic decomposition & 66.5 & 76.6 & 57.4 & 76.3 & 87.2 & 71.0 & 72.3 & 55.1 \\
 & atomic + re-score    & 68.8 & 70.9 & 57.9 & 66.0 & 88.1 & 72.2 & 75.5 & 51.2 \\
\midrule
\multirow{4}{*}{Llama3.1-8B}
 & Direct               & 76.4 & 34.9 & 67.4 & 29.7 & 81.7 & 47.8 & 67.6 & 29.9 \\
 & naive 5W1H           & 69.3 & 66.4 & 51.4 & 72.5 & 69.0 & 59.5 & 56.4 & 55.6 \\
 & atomic decomposition & 71.5 & 71.1 & 60.0 & 64.7 & 88.1 & 67.4 & 73.4 & 48.8 \\
 & atomic + re-score    & 71.7 & 62.7 & 57.5 & 53.7 & 87.1 & 65.7 & 80.3 & 47.4 \\
\midrule
\multirow{4}{*}{Gemma2-9B}
 & Direct               & 74.8 & 44.8 & 61.9 & 37.2 & 88.7 & 49.7 & 76.1 & 40.0 \\
 & naive 5W1H           & 66.2 & 58.7 & 66.3 & 54.5 & 61.1 & 50.5 & 68.6 & 40.9 \\
 & atomic decomposition & 76.5 & 68.1 & 64.2 & 62.4 & 93.3 & 68.1 & 73.9 & 52.7 \\
 & atomic + re-score    & 75.3 & 62.2 & 67.9 & 59.6 & 94.4 & 62.2 & 84.0 & 47.4 \\
\bottomrule
\end{tabular}
}
\caption{\textbf{Atomic-decomposition robustness check (full grid).} The second decomposition scheme across all primary backbones. \emph{Direct} and \emph{naive 5W1H} are repeated from Table~\ref{tab:decomp-repair}. Atomic decomposition does not reproduce the largest naive-5W1H losses (worst regression vs.\ Direct $-7.4$, against $-27.6$ for naive 5W1H). On an already-adaptive scheme, claim-conditioned re-scoring behaves as an operating-point shift: accuracy edges up in most cells while WSP declines in 11 of 12.}
\label{tab:atomic-app}
\end{table}

\end{document}